\documentclass[10pt,twocolumn,letterpaper]{article}

\usepackage{cvpr}              

\usepackage{booktabs}
\usepackage{multirow}
\usepackage{dblfloatfix}

\usepackage{float}

\usepackage{microtype}

\definecolor{cvprblue}{rgb}{0.21,0.49,0.74}
\usepackage[pagebackref,breaklinks,colorlinks,allcolors=cvprblue]{hyperref}
\usepackage{graphicx, booktabs, multirow, colortbl, xcolor, array, pifont, adjustbox, makecell}
\usepackage{tikz}
\usepackage{graphicx}
\usepackage{tcolorbox}
\usepackage{booktabs}
\usepackage{varwidth}
\usetikzlibrary{positioning, arrows.meta, shapes, fit}
\usepackage{tikz}
\usepackage{tikz}
\usepackage{pgfplots}
\pgfplotsset{compat=1.18}
\usepgfplotslibrary{polar}
\usepackage{graphicx}
\usetikzlibrary{positioning, arrows.meta, shapes, fit}
\usepackage{tikz}
\usepackage{graphicx}
\usetikzlibrary{positioning, arrows.meta}
\usepackage{booktabs, colortbl, xcolor, graphicx, multirow, adjustbox, array}

\def\paperID{*****} 
\def\confName{EarthVision}
\def\confYear{2026}

\title{
Where Do Vision-Language Models Fail? \\
World Scale Analysis for Image Geolocalization}

\author{Siddhant Bharadwaj$^{1}$  \hspace{1cm} Ashish Vashist$^{2}$  \hspace{1cm} Fahimul Aleem$^{3}$ \hspace{1cm} Shruti Vyas$^{4}$\\
University of Central Florida\\
United States\\
{\tt\small $^{1}$siddhant0701@gmail.com, $^{2}$avashist@andrew.cmu.edu, $^{3}$fahimul.aleem@ucf.edu, $^{4}$shruti@ucf.edu}
}

\begin{document}
\maketitle
\begin{abstract}
Image geolocalization has traditionally been addressed through retrieval-based place recognition or geometry-based visual localization pipelines. Recent advances in Vision-Language Models (VLMs) have demonstrated strong zero-shot reasoning capabilities across multimodal tasks, yet their performance in geographic inference remains underexplored. In this work, we present a systematic evaluation of multiple state-of-the-art VLMs for country-level image geolocalization using ground-view imagery only. Instead of relying on image matching, GPS metadata, or task-specific training, we evaluate prompt-based country prediction in a zero-shot setting. The selected models are tested on three geographically diverse datasets to assess their robustness and generalization ability. Our results reveal substantial variation across models, highlighting the potential of semantic reasoning for coarse geolocalization and the limitations of current VLMs in capturing fine-grained geographic cues. This study provides the first focused comparison of modern VLMs for country-level geolocalization and establishes a foundation for future research at the intersection of multimodal reasoning and geographic understanding. The code is available at \url{https://github.com/Fahim17/EarthVision26_55.git}
\end{abstract}    
\vspace{-10pt}
\section{Introduction}
\label{sec:intro}


Image geolocalization aims to determine the geographic location where a query image was captured. Depending on the required granularity, the task ranges from GPS coordinates\cite{arandjelovic2016netvlad, berton2021adaptive, ge2020self, jin2017learned, liu2021densernet} to coarse grained localization such as city or country level recognition\cite{ qian2025earth}. Traditionally, visual localization and visual place recognition have relied on geometric reasoning and image retrieval pipelines built upon handcrafted or learned visual features\cite{ vo2016localizing, cai2019ground, workman2015location}. Retrieval-based methods, in contrast, approximate location by matching a query image against a geo-tagged image database and transferring the pose or coordinates of the most similar references\cite{zheng2025geox, shetty2019uav, zhang2023cross, li2024unleashing}. While these approaches have demonstrated strong performance in controlled settings, they often require large, curated databases, careful engineering choices, and computationally intensive matching procedures.

Recent advances in vision foundation models have reshaped the landscape of visual understanding. Vision-Language Models (VLMs)\cite{radford2021learning, chen2024internvl, bai2308qwen, liu2023visual, li2023blip}, pretrained on large-scale image-text pairs, exhibit strong zero-shot and few-shot generalization across diverse tasks, including classification, captioning, and reasoning. Unlike traditional geolocalization pipelines that rely on explicit feature matching or metric learning, VLMs implicitly encode high-level semantic, cultural, architectural, and environmental cues that can be indicative of geographic context. This capability opens a new paradigm for image geolocalization: instead of retrieving the nearest visual neighbor, a model can directly infer the likely country of origin through prompt-based reasoning.

\begin{figure*}[t]
    \centering
    \includegraphics[width=\textwidth]{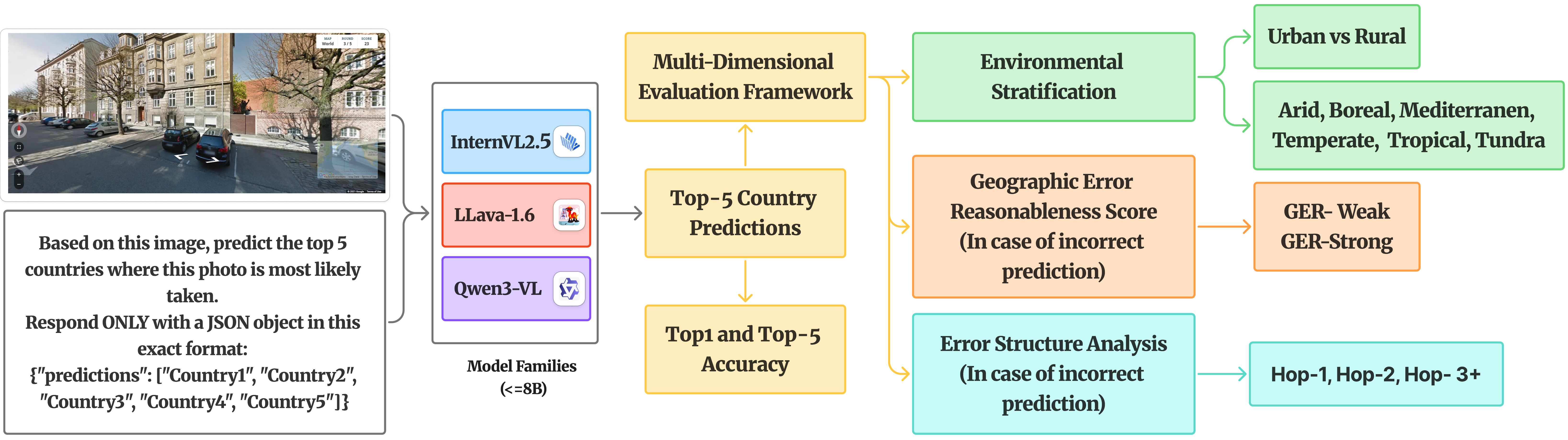}
    \caption{\textbf{Overview of the evaluation pipeline.} A ground-level image and a structured prompt are passed to each of the three VLM families (\(\leq\)8B parameters), which output Top-5 country predictions. Performance is assessed via Top-1/Top-5 accuracy and a 
multi-dimensional evaluation framework comprising Environmental 
Stratification (urban/rural and six biome categories), Error Structure 
Analysis (neighbor hop distance), and Geographic Error Reasonableness 
(GER) score, the latter two applied only to incorrect predictions.}
    \label{fig:your_label}
\end{figure*}

Although multimodal language models have been extensively evaluated on tasks such as visual question answering, captioning, and general visual reasoning, their ability to infer geographic location from images remains insufficiently explored. Furthermore, important aspects such as spatial reasoning capacity, sensitivity to cultural and environmental cues, and potential geographic biases are rarely analyzed in depth. Motivated by these gaps, we conduct a systematic and up-to-date benchmarking study of modern VLMs for country-level image geolocalization, providing both training-free (prompt-based) and controlled evaluation settings to better understand their intrinsic geospatial reasoning capabilities. 

In this work, we investigate the country-level geolocalization ability of multiple state-of-the-art Vision-Language Models through a unified benchmarking framework. We evaluate several VLMs on three ground-image-only datasets, where each query image is processed with a carefully designed prompt. By focusing exclusively on ground-view images and eliminating cross-view matching components, our study isolates the geographic reasoning capacity embedded within pretrained multimodal representations. We provide a comprehensive comparison across models, analyze performance trends, and examine how prompt design influences prediction behavior.

Figure~\ref{fig:your_label} provides an overview of the evaluation pipeline.

\noindent
Our benchmark offers two main contributions. 

\begin{enumerate}
    \item We provide a standardized evaluation protocol for prompt-based geolocalization using VLMs, reducing confounding factors related to training and architectural modifications.
    \item We introduce Geographic Error Reasonableness (GER) score, a novel metric that evaluates whether incorrect predictions are visually justified, and reveal key failure patterns including inverted scaling in the Qwen3-VL family, a consistent urban--rural performance gap across all models, and biome-dependent difficulty that varies across datasets.
\end{enumerate}

\section{Related Work}
\label{sec:related_work}
\textbf{Image Geolocalization and Place Recognition.}
Image Geolocalization, Place Recognition, and Landmark Retrieval are closely related tasks that associate images with geographic locations. Early approaches to image retrieval and landmark recognition relied on handcrafted local features such as SIFT and their aggregation using Fisher Vectors\cite{perronnin2007fisher} and VLAD\cite{jegou2011aggregating}. The introduction of deep convolutional neural networks (CNNs) significantly improved performance by learning discriminative global descriptors. Architectures such as NetVLAD\cite{arandjelovic2016netvlad} enabled end-to-end training for place recognition by integrating differentiable VLAD pooling into CNN backbones. Alternative pooling strategies including GeM\cite{philbin2007object} and R-MAC\cite{gordo2017end} further improved compactness and retrieval accuracy.
Traditional assessments focus on analyzing feature representations, descriptor compactness, and retrieval accuracy under large-scale settings\cite{berton2022deep, pion2020benchmarking}. However, these studies are typically designed for image matching pipelines rather than direct semantic reasoning about geographic attributes.

\noindent
\textbf{VLMs for Geo-Localization.}
More recently, Vision Transformers (ViT) have gained prominence for their ability to capture long-range global dependencies and explicitly encode positional information. For example, SemGeo \cite{rodrigues2023semgeo} improves retrieval in FoV-constrained scenarios by incorporating semantic cues from image segmentation and OpenStreetMap (OSM) data. Extending beyond static images, GAMa-Net \cite{vyas2022gama} tackles cross-view video geo-localization by leveraging temporal context from ground-level videos. Ye et al. \cite{ye2025cross} further broaden the scope by introducing natural language-guided retrieval, using Large Multimodal Models (LMMs) to generate detailed scene descriptions for the CVG-Text dataset and proposing CrossText2Loc for long-text-based localization. At the same time, the paradigm of image geolocalization is shifting from traditional retrieval-based pipelines to generative, reasoning-driven approaches powered by advanced Vision-Language Models (VLMs) such as GPT-4o\cite{hurst2024gpt}, Gemini 1.5\cite{team2024gemini}, and LLaVA\cite{liu2023visual}. These models rely on internal world knowledge to infer location from cues like architecture, vegetation, and infrastructure, without requiring explicit reference databases \cite{mu2026globalgeotree, tao2025advancements}. To systematically evaluate these capabilities, new benchmarks have emerged such as GRE-Suite\cite{wang2025gre} which analyzes reasoning chains for localization.

\noindent
In contrast to retrieval-based geolocalization methods, our work conducts a systematic evaluation of multiple SOTA VLMs under a unified prompting framework. We focus exclusively on ground-view images and assess zero-shot geographic reasoning performance across three datasets. This setting isolates the intrinsic geographic knowledge and multimodal reasoning capabilities of modern VLMs, providing complementary insights into traditional retrieval-based and metric-learning approaches. The closest prior work~\cite{roberts2024charting} evaluates GPT-4V on 100 images from GeoGuessr-50k the same dataset we use across four models; our benchmark scales this to nine models and three datasets, introducing structured error metrics including GER and hop distance analysis.

\section{Evaluation Protocol}
\label{sec:eval_protocol}
\subsection{Task Formulation}\label{sec:task_formulation}

We evaluate geographic localization under two complementary inference settings: unconstrained prediction and label-constrained prediction.

\noindent
\textbf{Unconstrained prediction.}
Given an input image, the model is prompted to predict the top five countries where the image is most likely taken. No candidate list is provided. The model generates a ranked list of five country names in structured JSON format:
\texttt{\{"predictions": ["Country1", ..., "Country5"]\}}
This setting evaluates geographic reasoning under free-form generation.

\noindent
\textbf{Label-constrained prediction.}
In the constrained setting, the model is provided with the complete set of valid country labels present in the evaluation dataset and is explicitly instructed to select predictions only from this predefined label space. The valid country set is dataset-specific and corresponds exactly to the ground-truth label space of the dataset being evaluated. This setting evaluates geographic discrimination under a fixed hypothesis space and reduces variability due to open-vocabulary generation errors.

\noindent
In both settings, models are required to output exactly five ranked country predictions in JSON format. Outputs that fail to conform to this structure or contain invalid country names are treated as incorrect. All experiments use greedy decoding (no sampling) to ensure deterministic evaluation.

\subsection{Prediction Normalization and Accuracy Metrics}\label{sec:metrics}

Model outputs are parsed as structured JSON objects containing a ranked list of five predicted countries. If parsing fails or the required key is missing, the prediction is treated as empty.
Predicted country names are normalized using case-insensitive exact string matching against the valid country label set for the dataset. Predictions not present in the valid label set are discarded. No fuzzy matching or alias expansion is applied.
Let $\mathcal{P}_k$ denote the top-$k$ valid predictions after filtering, and let $y$ denote the ground-truth country.

\textbf{Top-1 Accuracy} is defined as the proportion of samples for which $y \in \mathcal{P}_1$.

\textbf{Top-5 Accuracy} is defined as the proportion of samples for which $y \in \mathcal{P}_5$.

If no valid predictions remain after normalization, the sample is counted as incorrect.

\subsection{Models Evaluated}\label{sec:models}

We evaluate nine multimodal vision-language models spanning three families: InternVL2.5, LLaVA-based variants, and Qwen3-VL. Model sizes range from 1B to 8B parameters.
All models are evaluated in their publicly released pretrained form without task-specific fine-tuning. No additional training or adaptation is performed.
Table~\ref{tab:models} compares the model architectures and shows parameter counts of all evaluated models.

For consistency across models, we use greedy decoding in all experiments. Image preprocessing follows each model's recommended inference pipeline. No model-specific prompt engineering beyond the structured JSON instruction described in Section~\ref{sec:task_formulation} is applied.

\begin{table}[t]
\centering
\caption{Vision-language models evaluated. All models are used in their publicly released instruct-tuned form without task-specific fine-tuning. [*L=LLaVA]}
\resizebox{\columnwidth}{!}{%
\begin{tabular}{llllc}
\toprule
\textbf{Model} & \textbf{Vision Encoder} & \textbf{Language Model} & \textbf{Resolution} \\
\midrule
InternVL2.5-1B \cite{chen2024internvl2_5} & InternViT-300M & Qwen2.5-0.5B & 448 (dyn.)  \\
InternVL2.5-4B \cite{chen2024internvl2_5} & InternViT-300M & Qwen2.5-3B & 448 (dyn.)  \\
InternVL2.5-8B \cite{chen2024internvl2_5} & InternViT-300M & InternLM2.5-7B & 448 (dyn.)  \\
\midrule
*L-Mistral-7B \cite{liu2024llavanext}  & CLIP ViT-L/14 & Mistral-7B-v0.2 & 336  \\
*L-Vicuna-7B \cite{liu2024llavanext} & CLIP ViT-L/14 & Vicuna-7B-v1.5 & 336  \\
*L3-LLaVA-8b \cite{liu2024llavanext}  & CLIP ViT-L/14 & LLaMA-3-8B & 336  \\
\midrule
Qwen3-VL-2B \cite{qwen3technicalreport} & Qwen3-ViT & Qwen3-2B & Dynamic  \\
Qwen3-VL-4B \cite{qwen3technicalreport} & Qwen3-ViT & Qwen3-4B & Dynamic  \\
Qwen3-VL-8B \cite{qwen3technicalreport} & Qwen3-ViT & Qwen3-8B & Dynamic  \\
\bottomrule
\end{tabular}%
}
\label{tab:models}
\end{table}

\subsection{Environmental Stratification}\label{sec:env_stratification}

To mitigate dependence on any single representation model, we perform environmental labelling using three independent annotators: \textbf{CLIP} (ViT-L/14), \textbf{SigLIP} (google/siglip-so400m-patch14-384), and \textbf{Qwen3-VL-4B-Instruct}.

\noindent
\textbf{Urban vs.\ Rural Classification.}
For CLIP and SigLIP, we perform zero-shot classification by computing cosine similarity between each image embedding and text embeddings of two prompts: \emph{``an urban city scene''} and \emph{``a rural countryside scene''}. The label with higher softmax-normalized similarity is assigned.
For Qwen3-VL-4B, we prompt the model directly to classify each image as urban or rural.
Results are reported as mean $\pm$ std across the three labellers.

\noindent
\textbf{Biome-Level Categorization.}
We define six biomes---\emph{Tropical}, \emph{Arid}, \emph{Temperate}, \emph{Mediterranean}, \emph{Tundra}, and \emph{Boreal}---each with a descriptive prompt (e.g., \emph{``a dry desert or arid landscape''}). (see Table~\ref{tab:biome_prompts} in the supplementary material)
For CLIP and SigLIP, each image is assigned the highest-scoring biome via softmax-normalized cosine similarity; for Qwen3-VL-4B, we prompt the model to select among the six categories.
To obtain high-confidence labels, we retain only samples where all three annotators assign the identical biome (\emph{consensus filtering}), yielding subsets of 13{,}400 (26.8\%) on GeoGuessr-50k, 10{,}264 (15.0\%) on CityGuessr, and 18{,}001 (36.0\%) on OSV5M. The distribution of consensus biome labels across datasets is provided in the supplementary material (Table~\ref{tab:suppl_biome_consensus_dist}).

\subsection{Error Structure Analysis}\label{sec:error_analysis}

\textbf{Neighbor Hop Distance.}
\label{sec:hop_method}
To assess geographic coherence of errors, we measure border-hop distance between predicted and ground-truth countries. We construct an undirected country adjacency graph from a world border dataset, including pass-through countries for realistic shortest paths. Special edges connect territories and enclaves (e.g., Hong Kong–China), and island nations are linked to their nearest neighbor via haversine distance. Remaining disconnected components are bridged to ensure full connectivity.
For each incorrect Top-1 prediction, we compute the shortest-path distance (number of border crossings) to the ground-truth country. We report the percentage of errors at Hop-1, Hop-2, and Hop-3. Correct predictions are excluded.

\subsection{Geographic Error Reasonableness (GER) Score}\label{sec:ger_method}

\begin{figure*}[t]
\centering
\includegraphics[width=\textwidth]{JER_fixed.png}
\caption{Illustrative examples of Geographic Error 
Reasonableness score (GER). \textbf{Top:} High-GER error --- 
Qwen3-VL-4B predicts Singapore for a Malaysian image; 
all five nearest CLIP neighbors are from Singapore, 
indicating a visually justified confusion between two 
countries with shared urban landscapes. \textbf{Bottom:} 
Low-GER error --- the same model predicts Brazil for a 
Malaysian image; none of the five neighbors match Brazil, 
exposing an arbitrary misclassification with no visual 
support. GER distinguishes these qualitatively different 
failure modes, which standard accuracy metrics conflate.}
\label{fig:ger_example}
\end{figure*}

Standard evaluation metrics for geolocalization---Top-$k$ 
accuracy and geographic distance---treat all incorrect 
predictions equally. A model that predicts Thailand for an 
image taken in Cambodia is penalized identically to one 
that predicts Canada. Yet any experienced GeoGuessr player 
recognizes that not all errors are equal: confusing two 
countries with similar street infrastructure, vegetation, 
and signage is a fundamentally different failure mode from 
producing a geographically and visually arbitrary guess. 
Hop distance (Section~\ref{sec:hop_method}) partially 
addresses this by measuring spatial proximity of errors, 
but geographic adjacency does not guarantee visual 
similarity---neighboring countries can look drastically 
different (e.g., Egypt and Sudan share a border but differ 
in urban landscape), while distant countries may share 
strong visual resemblance (e.g., Portugal and Brazil).

We introduce \emph{Geographic Error Reasonableness (GER) score}, 
a metric that evaluates whether an incorrect prediction is 
\emph{visually justified} by the local appearance manifold 
of the dataset. The key intuition is that a wrong answer 
should not be penalized as harshly if the query image 
genuinely looks like it could have come from the predicted 
country---precisely the kind of ``reasonable mistake'' that 
human players routinely make. 

\noindent
\textbf{Definition.}
Let $\mathcal{D} = \{(x_i, y_i)\}_{i=1}^{N}$ denote the 
evaluation dataset, where $x_i$ is an image and $y_i$ is 
the ground-truth country label. We compute image embeddings 
and define the visual neighborhood $\mathcal{N}_K(x_i)$ as 
the set of ground-truth country labels of the $K$ nearest 
images to $x_i$ in embedding space (excluding $x_i$ itself), 
using cosine similarity over $\ell_2$-normalized embeddings. 
We set $K{=}5$ throughout. To ensure robustness to the 
choice of encoder, we compute neighbors independently using 
both CLIP ViT-L/14 and SigLIP embeddings, and report 
mean $\pm$ std across the two.

For an incorrect Top-1 prediction $\hat{y}_i \neq y_i$, 
let $c(\hat{y}_i, x_i)$ denote the number of neighbors in 
$\mathcal{N}_K(x_i)$ whose ground-truth country matches 
$\hat{y}_i$:
\begin{equation}
c(\hat{y}_i, x_i) = \bigl|\{j : y_j = \hat{y}_i,\; 
x_j \in \mathcal{N}_K(x_i)\}\bigr|
\label{eq:ger_count}
\end{equation}

We define two thresholds of visual justification:
\begin{equation}
\text{GER-Weak}  = \frac{1}{|\mathcal{E}|} 
\sum_{i \in \mathcal{E}} 
\mathbf{1}\bigl[c(\hat{y}_i, x_i) \geq 1\bigr]
\label{eq:ger_weak}
\end{equation}
\begin{equation}
\text{GER-Strong} = \frac{1}{|\mathcal{E}|} 
\sum_{i \in \mathcal{E}} 
\mathbf{1}\bigl[c(\hat{y}_i, x_i) \geq \tau\bigr], 
\quad \tau = 2
\label{eq:ger_strong}
\end{equation}
where $\mathcal{E} = \{i : \hat{y}_i \neq y_i\}$ is the 
set of incorrectly predicted samples.

GER-Weak measures the proportion of errors where the 
predicted country appears among at least one visual 
neighbor, indicating minimal visual plausibility. 
GER-Strong requires the predicted country to appear in at 
least $\tau{=}2$ of the five neighbors ($\geq$40\% of the 
neighborhood), indicating strong visual support for the 
confusion. Both are reported as percentages over incorrect 
predictions only.

Figure~\ref{fig:ger_example} provides a concrete example: 
when Qwen3-VL-4B predicts Singapore for a Malaysian image, 
all five CLIP neighbors originate from Singapore 
($c = 5$), yielding high GER; when it predicts Brazil, 
no neighbors match ($c = 0$), yielding zero GER.

\noindent
\textbf{Interpretation.}
A high GER score indicates that a model's errors are 
concentrated in visually ambiguous regions of the dataset---the model is ``wrong for the right reasons,'' much like 
a GeoGuessr player who guesses a plausible but ultimately 
incorrect neighboring country. A low GER score suggests 
that errors are visually arbitrary, reflecting poor 
geographic grounding rather than genuine visual ambiguity. 
Crucially, GER complements hop distance: hop distance 
measures \emph{where} a model errs geographically, while 
GER measures \emph{why} it errs visually.
\section{Datasets}
\label{sec:datasets}

We evaluate on three geographically diverse benchmarks spanning different image sources, geographic scales, and label spaces. Table~\ref{tab:dataset_summary} summarizes key statistics. All datasets are used exclusively for evaluation; no training or fine-tuning is performed.

\begin{table}[h]
\centering
\caption{\textbf{Summary of evaluation datasets.} $^\dagger$GSV=Google Street View. $^\ddagger$Stratified subset of the full test split.
}
\label{tab:dataset_summary}
\begin{tabular}{lcccl}
\toprule
\textbf{Dataset} & \textbf{Images} & \textbf{Countries} & \textbf{Source} \\
\midrule
GeoGuessr50k \cite{geoguessr50k} & 49{,}997 & 124 & $^\dagger$GSV  \\
CityGuessr68k \cite{cityguessr} & 68{,}269 & 91 & YouTube  \\
OSV5M$^\ddagger$ \cite{osv5m} & 50{,}000 & 219 & Mapillary  \\
\bottomrule
\end{tabular}%
\end{table}

\textbf{GeoGuessr-50k}~\cite{geoguessr50k} contains approximately 50{,}000 Google Street View screenshots collected from the GeoGuessr online game, each annotated with a country-level label. The dataset spans 124 countries but is heavily imbalanced: the United States alone accounts for roughly one-fifth of all images. Images retain the game interface overlay (compass, map widget), which may introduce visual artifacts. We use the full dataset (49{,}997 images) for evaluation.

\textbf{CityGuessr68k}~\cite{cityguessr} is a global-scale video geolocalization  dataset introduced at ECCV~2024, containing 68{,}269 YouTube-sourced videos from 166 cities across 91 countries, with faces blurred using RetinaFace~\cite{retinaface}. The dataset provides hierarchical labels at city, state, country, and continent levels; we use only the country-level annotation. Since our evaluation targets static image geolocalization , we extract a single representative frame per video by selecting the \textbf{middle frame} of each sequence.

\textbf{OpenStreetView-5M (OSV5M)}~\cite{osv5m} is a large-scale, open-access dataset of over 5.1 million geo-referenced street-view images sourced from Mapillary, covering 225 countries and territories. Introduced at CVPR~2024, it enforces a strict train/test separation. Due to the large test set size, we construct a \textbf{stratified evaluation subset} of 50{,}000 images by guaranteeing at least 20 images per country and distributing the remaining samples proportionally, ensuring both minimum representation and preservation of the overall geographic distribution. The resulting subset spans 219 countries and territories.

The three benchmarks provide complementary evaluation axes: GeoGuessr-50k covers 124 countries with a Western/developed-country bias, CityGuessr68k spans 91 countries with emphasis on major cities, and OSV5M provides the broadest coverage at 219 countries. The number of valid country labels varies per dataset, directly affecting task difficulty under both unconstrained and label-constrained prediction (Section~\ref{sec:task_formulation}).

\section{Results}

We evaluate nine vision-language models (Section~\ref{sec:models}) across three geolocation benchmarks---GeoGuessr-50k, CityGuessr, and OpenStreetView5M (OSV5M)---for country-level prediction and fine-grained error analysis.
Results are organized into five dimensions: (i)~overall accuracy under both unconstrained and label-constrained settings (Section~\ref{sec:task_formulation}), (ii)~urban--rural stratification validated across multiple labellers, (iii)~biome-level breakdown using consensus labels (Section~\ref{sec:env_stratification}), (iv)~structured error analysis via Neighbor Hop Distance (Section~\ref{sec:error_analysis}) and Geographic Error Reasonableness score (Section~\ref{sec:ger_method}), and (v)~preliminary LoRA fine-tuning.

\subsection{Overall Geolocalization Accuracy}

\definecolor{lvl0}{RGB}{255,255,255}   
\definecolor{lvl1}{RGB}{224,242,224}   
\definecolor{lvl2}{RGB}{169,217,169}   
\definecolor{lvl3}{RGB}{99,183,99}     
\definecolor{lvl4}{RGB}{34,139,34}     

\begin{table}[h]
\centering
\caption{
  \textbf{Geolocalization accuracy.} Top-1 / Top-5 accuracy (\%) across all three benchmarks.
  \textbf{Bold} = best results. {[*L = LLaVA]}.
  Cell shading:
  \colorbox{lvl0}{\small $<$40}\ 
  \colorbox{lvl1}{\small 40--50}\ 
  \colorbox{lvl2}{\small 50--60}\ 
  \colorbox{lvl3}{\small 60--70}\ 
  \colorbox{lvl4}{\textcolor{white}{\small 70+}}.
}
\label{tab:top1_top5}
\setlength{\tabcolsep}{5pt}
\renewcommand{\arraystretch}{1.22}

\begin{adjustbox}{max width=\columnwidth}
\begin{tabular}{l cc cc cc}
\toprule
& \multicolumn{2}{c}{\textbf{GeoGuessr}} 
& \multicolumn{2}{c}{\textbf{CityGuessr}} 
& \multicolumn{2}{c}{\textbf{OSV5M}} \\
\cmidrule(lr){2-3}\cmidrule(lr){4-5}\cmidrule(lr){6-7}
\textbf{Model} & \textbf{Top-1} & \textbf{Top-5} & \textbf{Top-1} & \textbf{Top-5} & \textbf{Top-1} & \textbf{Top-5} \\
\midrule

InternVL2.5-1B
  & \cellcolor{lvl0}36.26
  & \cellcolor{lvl3}66.51
  & \cellcolor{lvl1}45.01
  & \cellcolor{lvl3}65.68
  & \cellcolor{lvl0}30.34
  & \cellcolor{lvl2}51.45 \\

InternVL2.5-4B
  & \cellcolor{lvl2}55.97
  & \cellcolor{lvl4}\textcolor{white}{74.40}
  & \cellcolor{lvl2}55.05
  & \cellcolor{lvl3}68.84
  & \cellcolor{lvl0}32.06
  & \cellcolor{lvl2}50.55 \\

InternVL2.5-8B
  & \cellcolor{lvl3}65.00
  & \cellcolor{lvl4}\textcolor{white}{82.64}
  & \cellcolor{lvl2}57.21
  & \cellcolor{lvl4}\textcolor{white}{70.52}
  & \cellcolor{lvl0}36.30
  & \cellcolor{lvl2}54.81 \\

\midrule

*L-Mistral-7B
  & \cellcolor{lvl0}39.42
  & \cellcolor{lvl2}53.69
  & \cellcolor{lvl0}30.51
  & \cellcolor{lvl1}46.49
  & \cellcolor{lvl0}17.67
  & \cellcolor{lvl0}35.57 \\

*L-Vicuna-7B
  & \cellcolor{lvl0}35.75
  & \cellcolor{lvl2}53.76
  & \cellcolor{lvl0}36.70
  & \cellcolor{lvl2}50.91
  & \cellcolor{lvl0}19.23
  & \cellcolor{lvl0}36.39 \\

*L3-LLaVA-8B
  & \cellcolor{lvl1}49.86
  & \cellcolor{lvl3}68.90
  & \cellcolor{lvl0}39.10
  & \cellcolor{lvl2}56.71
  & \cellcolor{lvl0}27.96
  & \cellcolor{lvl1}46.71 \\

\midrule

Qwen3-VL-2B
  & \cellcolor{lvl4}\textcolor{white}{73.86}
  & \cellcolor{lvl4}\textcolor{white}{86.71}
  & \cellcolor{lvl3}64.57
  & \cellcolor{lvl4}\textcolor{white}{77.25}
  & \cellcolor{lvl1}\textbf{46.50}
  & \cellcolor{lvl3}64.48 \\

Qwen3-VL-4B
  & \cellcolor{lvl4}\textcolor{white}{\textbf{74.79}}
  & \cellcolor{lvl4}\textcolor{white}{\textbf{88.48}}
  & \cellcolor{lvl3}\textbf{65.78}
  & \cellcolor{lvl4}\textcolor{white}{78.05}
  & \cellcolor{lvl1}45.45
  & \cellcolor{lvl4}\textcolor{white}{\textbf{64.92}} \\

Qwen3-VL-8B
  & \cellcolor{lvl3}67.30
  & \cellcolor{lvl4}\textcolor{white}{80.19}
  & \cellcolor{lvl3}65.31
  & \cellcolor{lvl4}\textcolor{white}{\textbf{78.44}}
  & \cellcolor{lvl1}41.78
  & \cellcolor{lvl3}62.85 \\

\bottomrule
\end{tabular}
\end{adjustbox}

\smallskip
\noindent\footnotesize\textit{*L = LLaVA. Models separated by family: InternVL (top), LLaVA (middle), Qwen3 (bottom).}

\end{table}

Table~\ref{tab:top1_top5} presents Top-1 and Top-5 accuracy (Section~\ref{sec:metrics}) under unconstrained prediction across all three benchmarks.
The Qwen3-VL family dominates across the board, with Qwen3-VL-4B achieving the highest Top-1 accuracy on both GeoGuessr-50k (74.79\%) and CityGuessr (65.78\%), while Qwen3-VL-2B leads on OSV5M (46.50\%).
A notable finding is \emph{inverted scaling} in Qwen3-VL: the 8B variant 
consistently underperforms the 4B and 2B models (e.g., $-$7.49\,pp on 
GeoGuessr-50k), suggesting increased parameter count does not guarantee 
improved spatial reasoning. Importantly, as described in Section~\ref{sec:models}, the Qwen3-VL variants share the same vision encoder and differ primarily in the scale of the underlying language model, indicating that the observed inverted scaling stems from differences in language decoding rather than visual representation.

In contrast, the InternVL2.5 series exhibits conventional monotonic scaling, with steady gains from 1B to 8B across all three benchmarks---on OSV5M, the 8B variant reaches 36.30\% Top-1, surpassing the 4B (32.06\%) and 1B (30.34\%) models.
The LLaVA variants trail significantly: LLaVA-Mistral-7B and LLaVA-Vicuna-7B achieve the lowest accuracies on all three datasets, with OSV5M Top-1 scores of only 17.67\% and 19.23\% respectively, indicating weak country-level geographic grounding.

\noindent\textbf{Comparison with supervised models.} Notably, our zero-shot Qwen3-VL-4B (65.96\% Top-1) approaches the performance of GAEA (69.6\%)~\cite{campos2025gaea} on the CityGuessr validation set, despite GAEA being explicitly fine-tuned on CityGuessr data, and outperforms other task-specific baselines including LLaVA-OneVision (55.1\%), InternVL2-8B (52.3\%), and GLM-4V-9B (48.1\%).

\subsection{Urban vs.\ Rural Stratification}

Using multi-annotator urban/rural labels (Section~\ref{sec:env_stratification}) and cross-encoder visual neighbors (Section~\ref{sec:ger_method}), Table~\ref{tab:stratified_combined} jointly reports urban/rural Top-1 accuracy and GER scores across all three benchmarks.

\begin{table*}[t]
\centering
\caption{Per-dataset stratified results.
\textbf{Urban/Rural}: Top-1 accuracy (\%) reported as mean across three labellers (max SD 6.66\,pp).
\textbf{GER}: predicted country in $\geq$1 (W) / $\geq$2 (S) of 5
nearest CLIP neighbours (\%), over incorrect Top-1 only (higher is better); reported as mean across CLIP and SigLIP (max SD 1.44\,pp).
\textbf{Constrained}: label-constrained Top-1 (T1) / Top-5 (T5) accuracy (\%).
Full tables with per-cell std in supplementary
(Tables~\ref{tab:urban_rural_multi_full},~\ref{tab:ger_multi_full}).
[*L\,=\,LLaVA]}
\label{tab:stratified_combined}
\setlength{\tabcolsep}{3pt}
\renewcommand{\arraystretch}{1.15}
\begin{adjustbox}{max width=\textwidth}
\begin{tabular}{l cc cc cc cc cc cc cc cc cc}
\toprule
& \multicolumn{6}{c}{\textbf{GeoGuessr-50k}}
& \multicolumn{6}{c}{\textbf{CityGuessr}}
& \multicolumn{6}{c}{\textbf{OSV5M}} \\
\cmidrule(lr){2-7} \cmidrule(lr){8-13} \cmidrule(lr){14-19}
& \multicolumn{2}{c}{Urban/Rural}
& \multicolumn{2}{c}{GER}
& \multicolumn{2}{c}{Constrained}
& \multicolumn{2}{c}{Urban/Rural}
& \multicolumn{2}{c}{GER}
& \multicolumn{2}{c}{Constrained}
& \multicolumn{2}{c}{Urban/Rural}
& \multicolumn{2}{c}{GER}
& \multicolumn{2}{c}{Constrained} \\
\cmidrule(lr){2-3} \cmidrule(lr){4-5} \cmidrule(lr){6-7}
\cmidrule(lr){8-9} \cmidrule(lr){10-11} \cmidrule(lr){12-13}
\cmidrule(lr){14-15} \cmidrule(lr){16-17} \cmidrule(lr){18-19}
Model
& Urb & Rur & W & S & T1 & T5
& Urb & Rur & W & S & T1 & T5
& Urb & Rur & W & S & T1 & T5 \\
\midrule
InternVL2.5-1B
  & 48.8 & 27.7 & 16.0 & 7.2 & 3.25 & 36.22
  & 46.9 & 30.1 & 5.3 & 2.0 & 29.85 & 42.86
  & 33.6 & 30.2 & 11.5 & 4.7 & 16.09 & 21.38 \\
InternVL2.5-4B
  & 69.3 & 46.7 & 21.3 & 11.4 & 31.35 & 66.00
  & 57.6 & 34.9 & 7.3 & 2.9 & 50.96 & 66.25
  & 36.4 & 31.7 & 11.5 & 5.0 & 28.28 & 44.30 \\
InternVL2.5-8B
  & 74.5 & 58.4 & 33.7 & 18.7 & 52.13 & 78.70
  & 59.6 & 38.2 & 7.3 & 2.9 & 53.74 & 75.29
  & 38.2 & 36.3 & 14.1 & 6.2 & 26.18 & 54.55 \\
\midrule
*L-Mistral-7B
  & 46.5 & 34.4 & 13.3 & 7.0 & 28.32 & 39.15
  & 32.0 & 19.2 & 3.2 & 1.2 & 25.04 & 34.24
  & 19.6 & 17.4 & 5.7 & 2.4 & 16.18 & 18.44 \\
*L-Vicuna-7B
  & 42.8 & 30.7 & 10.7 & 5.7 & 31.11 & 38.54
  & 38.4 & 23.8 & 3.1 & 1.2 & 27.09 & 33.39
  & 22.3 & 18.8 & 5.5 & 2.5 & 3.91 & 5.70 \\
*L3-LLaVA-8B
  & 59.9 & 42.9 & 21.4 & 10.8 & 19.32 & 55.24
  & 40.7 & 26.1 & 4.0 & 1.4 & 33.59 & 49.63
  & 28.5 & 28.0 & 10.0 & 4.1 & 15.44 & 24.98 \\
\midrule
Qwen3-VL-2B
  & 82.6 & 67.8 & 40.5 & 23.2 & 61.40 & 76.75
  & 67.2 & 43.3 & 7.8 & 3.1 & 61.13 & 75.43
  & 51.0 & 46.1 & 17.2 & 7.6 & 29.91 & 42.70 \\
Qwen3-VL-4B
  & \textbf{83.8} & \textbf{68.5} & \textbf{44.4} & \textbf{25.9} & \textbf{74.69} & \textbf{89.78}
  & \textbf{68.8} & 41.5 & \textbf{8.3} & \textbf{3.3} & 70.46 & 84.97
  & \textbf{51.3} & 44.8 & \textbf{17.1} & \textbf{7.8} & \textbf{42.81} & 60.26 \\
Qwen3-VL-8B
  & 79.8 & 58.6 & 30.2 & 17.0 & 70.58 & 86.56
  & 68.3 & 41.8 & 7.8 & 3.1 & \textbf{70.52} & \textbf{85.71}
  & 48.2 & 41.0 & 15.4 & 6.7 & 42.42 & \textbf{61.22} \\
\bottomrule
\end{tabular}
\end{adjustbox}
\end{table*}

All models exhibit a substantial urban bias (Table~\ref{tab:stratified_combined}), consistently performing better on scenes with man-made landmarks, signage, and dense built environments.
This pattern is robust across all three labellers: on GeoGuessr-50k, Qwen3-VL-4B achieves 83.8\% urban versus 68.5\% rural Top-1, yielding a gap of ${\sim}$15\,pp regardless of which encoder defines the urban/rural boundary.
Urban accuracy is highly stable across labellers (std $<$ 1\,pp for most models), while rural accuracy exhibits higher variance (std up to 5.21\,pp for Qwen3-VL-4B on CityGuessr), suggesting the three encoders largely agree on what constitutes an urban scene but diverge on the rural boundary---consistent with the visual heterogeneity of rural imagery worldwide.

The urban--rural disparity is most pronounced on CityGuessr, where Qwen3-VL-4B achieves 68.84$\pm$0.78\% urban but only 41.54$\pm$5.21\% rural Top-1.
This likely reflects CityGuessr's larger label space and extreme class 
imbalance (91.2\% urban, ${\sim}$6k rural samples).

On OSV5M, the gap compresses markedly: Qwen3-VL-2B shows only a ${\sim}$5\,pp Top-1 gap (50.96$\pm$6.23\% urban vs.\ 46.07$\pm$1.14\% rural), and several LLaVA variants exhibit near-zero gaps (e.g., LLaMA3-LLaVA-8B: 28.53$\pm$5.17\% vs.\ 28.00$\pm$1.16\%).
The high standard deviations on OSV5M urban accuracy ($>$5\,pp) further confirm that urban/rural boundaries are least consistent across labellers on this dataset, likely due to OSV5M's Mapillary-sourced imagery which often captures ambiguous peri-urban scenes.

\subsection{Biome-Level Performance}
Using consensus-filtered biome labels (Section~\ref{sec:env_stratification}), Table~\ref{tab:biome_consensus} reports accuracy on high-confidence subsets where all three annotators agree: 13{,}400 samples (26.8\%) on GeoGuessr-50k, 10{,}264 (15.0\%) on CityGuessr, and 18{,}001 (36.0\%) on OSV5M.
 
\begin{table*}[t]
\centering
\caption{ \textbf{Biome.} Top-1 accuracy (\%) across biome categories using consensus labels
(all three annotators agree). Best per-biome results are \textbf{bold}.
Top-5 results are reported in Table~\ref{tab:biome_consensus_full}
(supplementary). [*L\,=\,LLaVA]}
\label{tab:biome_consensus}
\setlength{\tabcolsep}{4pt}
\renewcommand{\arraystretch}{1.15}
\begin{adjustbox}{width=\textwidth}
\begin{tabular}{l cccccc cccccc cccccc}
\toprule
& \multicolumn{6}{c}{\textbf{GeoGuessr-50k} ($n{=}$13{,}400)}
& \multicolumn{6}{c}{\textbf{CityGuessr} ($n{=}$10{,}264)}
& \multicolumn{6}{c}{\textbf{OSV5M} ($n{=}$18{,}001)} \\
\cmidrule(lr){2-7} \cmidrule(lr){8-13} \cmidrule(lr){14-19}
Model & Arid & Bor. & Med. & Tmp. & Tro. & Tun.
      & Arid & Bor. & Med. & Tmp. & Tro. & Tun.
      & Arid & Bor. & Med. & Tmp. & Tro. & Tun. \\
\midrule
InternVL2.5-1B
  & 35.8 & 11.0 & 40.2 & 20.1 & 11.5 & 22.9
  & 45.3 & 15.2 & 60.5 & 41.3 & 47.9 & 29.4
  & 26.8 & 60.5 & 14.6 & 38.6 & 12.0 & 30.1 \\
InternVL2.5-4B
  & 46.6 & 15.4 & 49.6 & 56.8 & 12.8 & 20.2
  & 52.7 & 37.4 & 70.6 & 50.4 & 47.9 & 38.8
  & 28.4 & 64.0 & 35.4 & 42.7 & 13.5 & 26.5 \\
InternVL2.5-8B
  & 51.6 & 48.0 & 56.3 & 66.5 & 25.6 & 41.3
  & 53.6 & 43.4 & 72.4 & 53.6 & 47.9 & 48.5
  & 30.3 & 70.7 & 42.7 & 45.9 & 12.0 & 42.7 \\
\midrule
*L-Mistral-7B
  & 32.3 & 16.4 & 28.1 & 51.0 &  7.7 & 20.2
  & 36.0 &  2.0 & 38.7 & 36.1 &  6.2 & 16.7
  & 20.4 &  4.7 &  6.2 & 28.6 &  3.2 & 28.5 \\
*L-Vicuna-7B
  & 33.2 & 12.9 & 30.3 & 48.4 &  2.6 & 22.9
  & 38.6 & 15.2 & 46.9 & 38.0 & 14.6 & 19.9
  & 22.4 &  3.3 & 14.6 & 30.9 &  2.5 & 24.3 \\
*L3-LLaVA-8B
  & 36.3 & 15.7 & 40.4 & 53.0 &  3.8 & 25.7
  & 36.7 & 27.3 & 48.9 & 38.2 & 20.8 & 36.5
  & 26.6 & 56.9 & 10.4 & 36.2 &  6.1 & 39.1 \\
\midrule
Qwen3-VL-2B
  & 60.1 & 43.1 & 72.9 & 74.3 & 41.0 & 48.6
  & 55.2 & 43.4 & 81.1 & 59.2 & \textbf{52.1} & 45.2
  & 38.3 & \textbf{76.8} & \textbf{59.4} & \textbf{56.4} & \textbf{20.5} & 52.1 \\
Qwen3-VL-4B
  & \textbf{60.5} & \textbf{52.1} & \textbf{73.8} & \textbf{74.6} & \textbf{43.6} & \textbf{52.3}
  & \textbf{60.4} & \textbf{46.5} & \textbf{81.9} & \textbf{59.5} & \textbf{52.1} & 53.3
  & \textbf{39.1} & 69.8 & 53.1 & 54.6 & 19.0 & \textbf{52.5} \\
Qwen3-VL-8B
  & 51.9 & 21.8 & 67.8 & 67.8 & 34.6 & 44.0
  & 57.1 & 34.3 & 78.7 & 58.4 & 45.8 & \textbf{57.0}
  & 35.4 & 62.0 & 56.2 & 52.3 & 18.3 & 52.4 \\
\bottomrule
\end{tabular}
\end{adjustbox}
\end{table*}

Table~\ref{tab:biome_consensus} reveals pronounced biome-dependent performance variation, validated on high-confidence consensus labels.
On GeoGuessr-50k, \textbf{Boreal} regions remain challenging even under consensus filtering: Qwen3-VL-4B achieves 52.1\% Top-1 on Boreal versus 73.8\% on Mediterranean.
OSV5M presents a \emph{reversed} ranking: \textbf{Tropical} is hardest (best: 20.5\%, Qwen3-VL-2B) while \textbf{Boreal} becomes easiest (best: 76.8\%, Qwen3-VL-2B).
This inversion persists under consensus labels, confirming it reflects the geographic distribution of the dataset---tropical regions span dozens of visually similar countries while boreal regions are concentrated in few northern countries with distinctive infrastructure---rather than a labelling artifact.

The inverted scaling pattern in Qwen3-VL is biome-specific and robust: on GeoGuessr-50k Boreal, the 8B variant (21.8\%) substantially trails the 4B (52.1\%) and 2B (43.1\%), and on OSV5M Boreal the gap is 76.8\% (2B) vs.\ 62.0\% (8B).
On CityGuessr, the 8B model partially recovers on Boreal (34.3\%) and Tundra (57.0\%).

\subsection{Structured Error Analysis}
\noindent
\textbf{Neighbor Hop Distance.}
\begin{table}[t]
\centering
\caption{\textbf{Hop Distance.} Error distribution by hop distance (\%). H-1 = adjacent country; H-3+ = 3 or more hops. [*L=LLaVA]}
\label{tab:hop_combined}
\setlength{\tabcolsep}{2pt}
\begin{adjustbox}{width=\columnwidth}
\begin{tabular}{l ccc ccc ccc}
\toprule
& \multicolumn{3}{c}{\textbf{GeoGuessr}} & \multicolumn{3}{c}{\textbf{CityGuessr}} & \multicolumn{3}{c}{\textbf{OSV5M}} \\
\cmidrule(lr){2-4} \cmidrule(lr){5-7} \cmidrule(lr){8-10}
Model & H-1 & H-2 & H-3+ & H-1 & H-2 & H-3+ & H-1 & H-2 & H-3+ \\
\midrule
InternVL2.5-1B  & 40.6 & 4.8  & 54.6 & 34.3 & 9.0  & 56.7 & 22.5 & 11.8 & 65.7 \\
InternVL2.5-4B  & 26.0 & 5.0  & 69.0 & 31.5 & 11.1 & 57.4 & 18.4 & 8.1  & 73.5 \\
InternVL2.5-8B  & 40.4 & 9.4  & 50.2 & 31.8 & \textbf{12.1} & 56.1 & 23.7 & 9.0  & 67.3 \\
*L-Mistral-7B   & 14.8 & 2.5  & 82.7 & 13.2 & 6.1  & 80.7 & 11.4 & 2.2  & 86.4 \\
*L-Vicuna-7B    & 8.5  & 1.0  & 90.5 & 10.2 & 3.0  & 86.8 & 11.2 & 1.6  & 87.2 \\
*L3-LLaVA-8b    & 25.4 & 7.4  & 67.2 & 23.6 & 9.2  & 67.2 & 18.6 & 9.0  & 72.4 \\
Qwen3-VL-2B     & 40.8 & 9.0  & 50.2 & \textbf{43.5} & 12.0 & 44.5 & \textbf{30.8} & \textbf{13.6} & \textbf{55.6} \\
Qwen3-VL-4B     & \textbf{42.4} & \textbf{9.8} & \textbf{47.8} & 42.9 & 11.6 & \textbf{45.5} & 29.8 & 12.4 & 57.8 \\
Qwen3-VL-8B     & 27.1 & 4.6  & 68.3 & 42.5 & 10.6 & 46.9 & 26.9 & 9.9  & 63.2 \\
\bottomrule
\end{tabular}
\end{adjustbox}
\end{table}
The hop distance distribution (Table~\ref{tab:hop_combined}), computed over the country adjacency graph described in Section~\ref{sec:hop_method}, provides insight into the \emph{geographic coherence} of errors across all three benchmarks.
Qwen3-VL-4B achieves the highest Hop-1 rate (42.4\%) and lowest Hop-3+ rate (47.8\%) on GeoGuessr-50k, indicating that when it errs, it typically confuses a country with its geographic neighbor---a far more reasonable failure mode than the distant errors characteristic of weaker models.

On CityGuessr, the Qwen3-VL family maintains this pattern, with all three variants concentrating ${\sim}$42--43\% of errors at Hop-1 and only ${\sim}$45--47\% at Hop-3+, while LLaVA-Vicuna-7B sends 86.8\% of its errors to Hop-3+.
Notably, the inverted scaling phenomenon is \emph{absent} in hop distributions on CityGuessr: the 2B, 4B, and 8B Qwen3-VL variants produce nearly identical Hop-1 rates (43.5\%, 42.9\%, 42.5\%), suggesting that geographic proximity of errors is robust to model scale even when raw accuracy is not.

On OSV5M, Qwen3-VL-2B leads in Hop-1 errors (30.8\%), while the gap between the 2B and 8B variants (30.8\% vs.\ 26.9\%) re-emerges, consistent with the inverted scaling pattern observed in accuracy.
Across all three benchmarks, there is a clear correlation between overall accuracy and the proportion of geographically proximate errors: models that perform better also fail more gracefully.

\noindent
\textbf{Geographic Error Reasonableness score (GER).} Using cross-encoder visual neighbors (Section~\ref{sec:ger_method}), Table~\ref{tab:stratified_combined} reports GER as mean $\pm$ SD across CLIP and SigLIP embeddings.

GER scores (Table~\ref{tab:stratified_combined}), computed using visual neighborhoods as defined in Section~\ref{sec:ger_method}, quantify whether incorrect predictions are supported by the local visual manifold---i.e., whether the predicted country appears among the ground-truth labels of an image's nearest neighbors.
Qwen3-VL-4B achieves the highest GER-Weak (44.4\%) and GER-Strong (25.9\%) on GeoGuessr-50k, meaning nearly half of its incorrect predictions correspond to countries that are visually indistinguishable from the query image's neighborhood.
This is more than quadruple the GER-Weak of LLaVA-Vicuna-7B (10.7\%), underscoring a qualitative difference in failure modes: stronger models confuse visually similar regions, while weaker models produce arbitrary misclassifications with no visual support.

Absolute GER scores drop markedly on CityGuessr (best GER-Weak: 8.3\%) and OSV5M (best GER-Weak: 17.1\%), reflecting the increased difficulty of country-level discrimination across more diverse geographic distributions.
Despite these lower absolute values, the relative ranking across model families is remarkably consistent: Qwen3-VL-4B leads on all three benchmarks, and the LLaVA variants consistently occupy the bottom tier (3.1--5.7\% GER-Weak).



Table~\ref{tab:stratified_combined} reports performance under label-constrained 
prediction (Section~\ref{sec:task_formulation}), where models select from a 
predefined set of valid country labels.
On GeoGuessr-50k, Qwen3-VL-4B leads with 74.69\% Top-1, closely matching its unconstrained result (74.79\%). On CityGuessr, Qwen3-VL-8B recovers to match Qwen3-VL-4B (70.52\% vs.\ 70.46\% Top-1), suggesting its weaker unconstrained performance stems from label generation difficulties rather than inferior visual understanding.
OSV5M reveals the opposite pattern: nearly all models degrade under constraints, with Qwen3-VL-2B dropping from 46.50\% to 29.91\%. We attribute this to OSV5M's broader country distribution and heavy rural bias (82.8\%), where models struggle without landmark-driven cues. The LLaVA variants show minimal benefit from label constraints across all benchmarks, indicating more fundamental limitations.

\noindent\textbf{Sensitivity to label ordering.} 
We observe that label-constrained performance exhibits sensitivity to the ordering of country names in the candidate list, consistent with prior findings on ordering effects~\cite{pezeshkpour2023order, tan2024ordermatters}. 
Systematic quantification is left to future work.

\noindent\textbf{Preliminary LoRA fine-tuning.}
Fine-tuning InternVL2.5-1B with LoRA yields a +39.04\,pp Top-1 gain on a 20-country subset; details are in the supplementary (Section~\ref{sec:suppl_lora}).


\subsection{Limitations} 
Our evaluation focuses on country-level predictions, and our exploration of fine-tuning remains preliminary. We also observe geographic imbalance in the datasets, with GeoGuessr-50k skewed toward the United States and OSV5M underrepresenting regions such as parts of Africa and Central Asia, which may impact performance.


\section{Conclusion}
\label{sec:conclusion}
This paper investigated the effectiveness of contemporary Vision-Language Models for country-level geolocalization using only ground-view images and prompt-based inference. By benchmarking nine VLMs across three datasets and validating environmental stratification across multiple independent annotators, we provided a unified evaluation framework that contrasts semantic reasoning approaches with traditional retrieval-based geolocalization pipelines. Our findings show that while VLMs demonstrate promising zero-shot geographic reasoning capabilities, their performance remains inconsistent and highly dependent on visual context and prompt formulation. Compared to classical retrieval-based methods, VLMs offer greater flexibility and scalability but currently lack the reliability required for fine-grained or high-precision localization tasks.

This work highlights the growing role of multimodal foundation models in geolocalization while identifying key challenges such as improving geographic grounding, prompt robustness, and hybrid methods that integrate semantic reasoning with retrieval techniques. It also suggests that exploring chain-of-thought prompting and related strategies may help determine whether current zero-shot limitations stem from model constraints or suboptimal instructions. Overall, combining semantic reasoning with lightweight task-specific adaptation appears to be a promising path toward narrowing the gap with fully supervised approaches.



{
    \small
    \bibliographystyle{ieeenat_fullname}
    \bibliography{main}

@String(CVPR= {IEEE Conf. Comput. Vis. Pattern Recog.})

@String(ECCV= {Eur. Conf. Comput. Vis.})

@String(ICASSP=	{ICASSP})

@String(AAAI = {AAAI})

@String(CVPR  = {CVPR})

@String(ECCV  = {ECCV})

@inproceedings{li2023blip,
  title={Blip-2: Bootstrapping language-image pre-training with frozen image encoders and large language models},
  author={Li, Junnan and Li, Dongxu and Savarese, Silvio and Hoi, Steven},
  booktitle={International conference on machine learning},
  pages={19730--19742},
  year={2023},
  organization={PMLR}
}

@article{liu2023visual,
  title={Visual instruction tuning},
  author={Liu, Haotian and Li, Chunyuan and Wu, Qingyang and Lee, Yong Jae},
  journal={Advances in neural information processing systems},
  volume={36},
  pages={34892--34916},
  year={2023}
}

@article{bai2308qwen,
  title={Qwen-VL: A versatile vision-language model for understanding, localization, text reading, and beyond. 2023. doi: 10. 48550},
  author={Bai, Jinze and Bai, Shuai and Yang, Shusheng and Wang, Shijie and Tan, Sinan and Wang, Peng and Lin, Junyang and Zhou, Chang and Zhou, Jingren},
  journal={arXiv preprint ARXIV.2308.12966}
}

@inproceedings{chen2024internvl,
  title={Internvl: Scaling up vision foundation models and aligning for generic visual-linguistic tasks},
  author={Chen, Zhe and Wu, Jiannan and Wang, Wenhai and Su, Weijie and Chen, Guo and Xing, Sen and Zhong, Muyan and Zhang, Qinglong and Zhu, Xizhou and Lu, Lewei and others},
  booktitle={Proceedings of the IEEE/CVF conference on computer vision and pattern recognition},
  pages={24185--24198},
  year={2024}
}

@inproceedings{radford2021learning,
  title={Learning transferable visual models from natural language supervision},
  author={Radford, Alec and Kim, Jong Wook and Hallacy, Chris and Ramesh, Aditya and Goh, Gabriel and Agarwal, Sandhini and Sastry, Girish and Askell, Amanda and Mishkin, Pamela and Clark, Jack and others},
  booktitle={International conference on machine learning},
  pages={8748--8763},
  year={2021},
  organization={PmLR}
}

@inproceedings{li2024unleashing,
  title={Unleashing unlabeled data: A paradigm for cross-view geo-localization},
  author={Li, Guopeng and Qian, Ming and Xia, Gui-Song},
  booktitle={Proceedings of the IEEE/CVF Conference on Computer Vision and Pattern Recognition},
  pages={16719--16729},
  year={2024}
}

@inproceedings{zhang2023cross,
  title={Cross-view image sequence geo-localization},
  author={Zhang, Xiaohan and Sultani, Waqas and Wshah, Safwan},
  booktitle={Proceedings of the IEEE/CVF Winter Conference on Applications of Computer Vision},
  pages={2914--2923},
  year={2023}
}

@inproceedings{shetty2019uav,
  title={Uav pose estimation using cross-view geolocalization with satellite imagery},
  author={Shetty, Akshay and Gao, Grace Xingxin},
  booktitle={2019 International Conference on Robotics and Automation (ICRA)},
  pages={1827--1833},
  year={2019},
  organization={IEEE}
}

@article{zheng2025geox,
  title={GeoX-Bench: Benchmarking Cross-View Geo-Localization and Pose Estimation Capabilities of Large Multimodal Models},
  author={Zheng, Yushuo and Ying, Jiangyong and Duan, Huiyu and Li, Chunyi and Zhang, Zicheng and Liu, Jing and Liu, Xiaohong and Zhai, Guangtao},
  journal={arXiv preprint arXiv:2511.13259},
  year={2025}
}

@inproceedings{workman2015location,
  title={On the location dependence of convolutional neural network features},
  author={Workman, Scott and Jacobs, Nathan},
  booktitle={Proceedings of the IEEE conference on computer vision and pattern recognition workshops},
  pages={70--78},
  year={2015}
}

@inproceedings{cai2019ground,
  title={Ground-to-aerial image geo-localization with a hard exemplar reweighting triplet loss},
  author={Cai, Sudong and Guo, Yulan and Khan, Salman and Hu, Jiwei and Wen, Gongjian},
  booktitle={Proceedings of the IEEE/CVF international conference on computer vision},
  pages={8391--8400},
  year={2019}
}

@inproceedings{vo2016localizing,
  title={Localizing and orienting street views using overhead imagery},
  author={Vo, Nam N and Hays, James},
  booktitle={European conference on computer vision},
  pages={494--509},
  year={2016},
  organization={Springer}
}

@inproceedings{berton2021adaptive,
  title={Adaptive-attentive geolocalization from few queries: A hybrid approach},
  author={Berton, Gabriele Moreno and Paolicelli, Valerio and Masone, Carlo and Caputo, Barbara},
  booktitle={Proceedings of the IEEE/CVF Winter Conference on Applications of Computer Vision},
  pages={2918--2927},
  year={2021}
}

@inproceedings{ge2020self,
  title={Self-supervising fine-grained region similarities for large-scale image localization},
  author={Ge, Yixiao and Wang, Haibo and Zhu, Feng and Zhao, Rui and Li, Hongsheng},
  booktitle={European conference on computer vision},
  pages={369--386},
  year={2020},
  organization={Springer}
}

@inproceedings{jin2017learned,
  title={Learned contextual feature reweighting for image geo-localization},
  author={Jin Kim, Hyo and Dunn, Enrique and Frahm, Jan-Michael},
  booktitle={Proceedings of the IEEE conference on computer vision and pattern recognition},
  pages={2136--2145},
  year={2017}
}

@inproceedings{liu2021densernet,
  title={Densernet: Weakly supervised visual localization using multi-scale feature aggregation},
  author={Liu, Dongfang and Cui, Yiming and Yan, Liqi and Mousas, Christos and Yang, Baijian and Chen, Yingjie},
  booktitle={Proceedings of the AAAI conference on artificial intelligence},
  volume={35},
  number={7},
  pages={6101--6109},
  year={2021}
}

@article{qian2025earth,
  title={Where on Earth? A Vision-Language Benchmark for Probing Model Geolocation Skills Across Scales},
  author={Qian, Zhaofang and Chen, Hardy and Wang, Zeyu and Zhang, Li and Wang, Zijun and Huang, Xiaoke and Liu, Hui and Tang, Xianfeng and Zheng, Zeyu and Tu, Haoqin and others},
  journal={arXiv preprint arXiv:2510.10880},
  year={2025}
}

@misc{geoguessr50k,
  author       = {Rohan K.},
  title        = {{GeoLocation -- GeoGuessr Images (50K)}},
  year         = {2021},
  howpublished = {Kaggle},
  url          = {https://www.kaggle.com/datasets/ubitquitin/geolocation-geoguessr-images-50k},
  note         = {Accessed: 2025}
}

@inproceedings{cityguessr,
  author    = {Kulkarni, Parth Parag and Nayak, Gourav Kumar and Shah, Mubarak},
  title     = {{CityGuessr: City-Level Video Geo-Localization on a Global Scale}},
  booktitle = {European Conference on Computer Vision (ECCV)},
  year      = {2024},
  pages     = {289--306},
  publisher = {Springer},
  doi       = {10.1007/978-3-031-73036-8_17}
}

@inproceedings{osv5m,
  author    = {Astruc, Guillaume and Dufour, Nicolas and Siglidis, Ioannis and Aronssohn, Constantin and Bouia, Nacim and Fu, Stephanie and Loiseau, Romain and Nguyen, Van Nguyen and Raude, Charles and Vincent, Elliot and Xu, Lintao and Zhou, Hongyu and Landrieu, Loic},
  title     = {{OpenStreetView-5M: The Many Roads to Global Visual Geolocation}},
  booktitle = {Proceedings of the IEEE/CVF Conference on Computer Vision and Pattern Recognition (CVPR)},
  year      = {2024}
}

@inproceedings{retinaface,
  author    = {Deng, Jiankang and Guo, Jia and Ververas, Evangelos and Kotsia, Irene and Zafeiriou, Stefanos},
  title     = {{RetinaFace: Single-Shot Multi-Level Face Localisation in the Wild}},
  booktitle = {Proceedings of the IEEE/CVF Conference on Computer Vision and Pattern Recognition (CVPR)},
  year      = {2020},
  pages     = {5203--5212}
}

@article{jegou2011aggregating,
  title={Aggregating local image descriptors into compact codes},
  author={J{\'e}gou, Herv{\'e} and Perronnin, Florent and Douze, Matthijs and S{\'a}nchez, Jorge and P{\'e}rez, Patrick and Schmid, Cordelia},
  journal={IEEE transactions on pattern analysis and machine intelligence},
  volume={34},
  number={9},
  pages={1704--1716},
  year={2011},
  publisher={IEEE}
}

@inproceedings{perronnin2007fisher,
  title={Fisher kernels on visual vocabularies for image categorization},
  author={Perronnin, Florent and Dance, Christopher},
  booktitle={2007 IEEE conference on computer vision and pattern recognition},
  pages={1--8},
  year={2007},
  organization={IEEE}
}

@inproceedings{arandjelovic2016netvlad,
  title={NetVLAD: CNN architecture for weakly supervised place recognition},
  author={Arandjelovic, Relja and Gronat, Petr and Torii, Akihiko and Pajdla, Tomas and Sivic, Josef},
  booktitle={Proceedings of the IEEE conference on computer vision and pattern recognition},
  pages={5297--5307},
  year={2016}
}

@inproceedings{philbin2007object,
  title={Object retrieval with large vocabularies and fast spatial matching},
  author={Philbin, James and Chum, Ondrej and Isard, Michael and Sivic, Josef and Zisserman, Andrew},
  booktitle={2007 IEEE conference on computer vision and pattern recognition},
  pages={1--8},
  year={2007},
  organization={IEEE}
}

@article{gordo2017end,
  title={End-to-end learning of deep visual representations for image retrieval},
  author={Gordo, Albert and Almazan, Jon and Revaud, Jerome and Larlus, Diane},
  journal={International Journal of Computer Vision},
  volume={124},
  number={2},
  pages={237--254},
  year={2017},
  publisher={Springer}
}

@inproceedings{berton2022deep,
  title={Deep visual geo-localization benchmark},
  author={Berton, Gabriele and Mereu, Riccardo and Trivigno, Gabriele and Masone, Carlo and Csurka, Gabriela and Sattler, Torsten and Caputo, Barbara},
  booktitle={Proceedings of the IEEE/CVF Conference on Computer Vision and Pattern Recognition},
  pages={5396--5407},
  year={2022}
}

@inproceedings{pion2020benchmarking,
  title={Benchmarking image retrieval for visual localization},
  author={Pion, No{\'e} and Humenberger, Martin and Csurka, Gabriela and Cabon, Yohann and Sattler, Torsten},
  booktitle={2020 International Conference on 3D Vision (3DV)},
  pages={483--494},
  year={2020},
  organization={IEEE}
}

@misc{campos2025gaea,
  title={GAEA: A Geolocation Aware Conversational Assistant},
  author={Ron Campos and Ashmal Vayani and Parth Parag Kulkarni and Rohit Gupta and Aritra Dutta and Mubarak Shah},
  year={2025},
  eprint={2503.16423},
  archivePrefix={arXiv},
  primaryClass={cs.CV},
  url={https://arxiv.org/abs/2503.16423}
}

@article{roberts2024charting,
  title={Charting New Territories: Exploring the Geographic 
         and Geospatial Capabilities of Multimodal LLMs},
  author={Roberts, Jonathan and L{\"u}ddecke, Timo and 
          Sheikh, Rehan and Han, Kai and Albanie, Samuel},
  journal={arXiv preprint arXiv:2311.14656},
  year={2024}
}

@misc{liu2024llavanext,
  title        = {LLaVA-NeXT: Improved reasoning, OCR, and world knowledge},
  author       = {Liu, Haotian and Li, Chunyuan and Li, Yuheng and Li, Bo and Zhang, Yuanhan and Shen, Sheng and Lee, Yong Jae},
  month        = {January},
  year         = {2024},
  url          = {https://llava-vl.github.io/blog/2024-01-30-llava-next/}
}

@misc{qwen3technicalreport,
      title={Qwen3 Technical Report}, 
      author={Qwen Team},
      year={2025},
      eprint={2505.09388},
      archivePrefix={arXiv},
      primaryClass={cs.CL},
      url={https://arxiv.org/abs/2505.09388}, 
}

@article{pezeshkpour2023order,
  title = {Large Language Models Sensitivity to The Order of Options in Multiple-Choice Questions},
  author = {Pezeshkpour, Pouya and Hruschka, Estevam},
  year = {2023},
  journal = {arXiv},
  url = {https://arxiv.org/abs/2308.11483}
}

@article{tan2024ordermatters,
  title = {Order Matters: Exploring Order Sensitivity in Multimodal Large Language Models},
  author = {Tan, Zhijie and Chu, Xu and Li, Weiping and Mo, Tong},
  year = {2024},
  journal = {arXiv},
  url = {https://arxiv.org/abs/2410.16983}
}

@article{chen2024internvl2_5,
  title={Expanding Performance Boundaries of Open-Source Multimodal Models with Model, Data, and Test-Time Scaling},
  author={Chen, Zhe and Wang, Weiyun and Cao, Yue and Liu, Yangzhou and Gao, Zhangwei and Cui, Erfei and Zhu, Jinguo and Ye, Shenglong and Tian, Hao and Liu, Zhaoyang and others},
  journal={arXiv preprint arXiv:2412.05271},
  year={2024},
  url={https://arxiv.org/abs/2412.05271}
}

@inproceedings{rodrigues2023semgeo,
  title={Semgeo: Semantic keywords for cross-view image geo-localization},
  author={Rodrigues, Royston and Tani, Masahiro},
  booktitle={ICASSP 2023-2023 IEEE International Conference on Acoustics, Speech and Signal Processing (ICASSP)},
  pages={1--5},
  year={2023},
  organization={IEEE}
}

@inproceedings{ye2025cross,
  title={Where am i? cross-view geo-localization with natural language descriptions},
  author={Ye, Junyan and Lin, Honglin and Ou, Leyan and Chen, Dairong and Wang, Zihao and Zhu, Qi and He, Conghui and Li, Weijia},
  booktitle={Proceedings of the IEEE/CVF International Conference on Computer Vision},
  pages={5890--5900},
  year={2025}
}

@inproceedings{vyas2022gama,
  title={Gama: Cross-view video geo-localization},
  author={Vyas, Shruti and Chen, Chen and Shah, Mubarak},
  booktitle={European Conference on Computer Vision},
  pages={440--456},
  year={2022},
  organization={Springer}
}

@article{hurst2024gpt,
  title={Gpt-4o system card},
  author={Hurst, Aaron and Lerer, Adam and Goucher, Adam P and Perelman, Adam and Ramesh, Aditya and Clark, Aidan and Ostrow, AJ and Welihinda, Akila and Hayes, Alan and Radford, Alec and others},
  journal={arXiv preprint arXiv:2410.21276},
  year={2024}
}

@article{team2024gemini,
  title={Gemini 1.5: Unlocking multimodal understanding across millions of tokens of context, 2024},
  author={Team, Gemini and Georgiev, Petko and Lei, Ving Ian and Burnell, Ryan and Bai, Libin and Gulati, Anmol and Tanzer, Garrett and Vincent, Damien and Pan, Zhufeng and Wang, Shibo and others},
  journal={URL https://arxiv. org/abs/2403.05530},
  volume={2403},
  year={2024}
}

@article{wang2025gre,
  title={Gre suite: Geo-localization inference via fine-tuned vision-language models and enhanced reasoning chains},
  author={Wang, Chun and Ye, Xiaojun and Pan, Xiaoran and Pan, Zihao and Wang, Haofan and Song, Yiren},
  journal={arXiv preprint arXiv:2505.18700},
  year={2025}
}

@article{mu2026globalgeotree,
  title={GlobalGeoTree: a multi-granular vision-language dataset for global tree species classification},
  author={Mu, Yang and Xiong, Zhitong and Wang, Yi and Shahzad, Muhammad and Essl, Franz and Kreft, Holger and van Kleunen, Mark and Zhu, Xiao Xiang},
  journal={Earth System Science Data},
  volume={18},
  number={2},
  pages={1379--1403},
  year={2026},
  publisher={Copernicus Publications G{\"o}ttingen, Germany}
}

@article{tao2025advancements,
  title={Advancements in vision--language models for remote sensing: Datasets, capabilities, and enhancement techniques},
  author={Tao, Lijie and Zhang, Haokui and Jing, Haizhao and Liu, Yu and Yan, Dawei and Wei, Guoting and Xue, Xizhe},
  journal={Remote Sensing},
  volume={17},
  number={1},
  pages={162},
  year={2025},
  publisher={MDPI}
}
}

\clearpage
\setcounter{section}{0}
\onecolumn

\newcommand{\cmark}{\ding{51}}
\newcommand{\xmark}{\ding{55}}

\definecolor{correct}{RGB}{198,239,206}
\definecolor{wrong}{RGB}{255,199,206}
\definecolor{partial}{RGB}{255,235,156}

\begin{center}
{\Large\textbf{Supplementary Material} \\ 
Where Do Vision-Language Models Fail? \\
World Scale Analysis for Image Geolocalization}
\end{center}

\noindent This supplementary provides additional implementation details, qualitative analysis and additional results. We show qualitative comparison of semantically similar images identified as confusing samples (Sec.~\ref{sec:suppl_qualitative}), biome label distributions and classification prompts (Secs.~\ref{sec:suppl_biome_dist}--\ref{sec:suppl_biome_prompts}), Top-5 biome accuracy (Sec.~\ref{sec:suppl_biome_top5}), preliminary LoRA fine-tuning (Sec.~\ref{sec:suppl_lora}), and full Urban/Rural and GER tables with per-cell standard deviations (Sec.~\ref{sec:suppl_full_tables}).

\section{Qualitative Analysis}
\label{sec:suppl_qualitative}


Table \ref{tab:neighbour_confusion} presents qualitative examples of neighboring-country confusion produced by \textbf{Qwen3-VL-4B-Instruct} under a blind two-turn prompting setup. Three image pairs are shown, each drawn from geographically close countries: (1) Pakistan vs. India, (2) Bangladesh vs. India, and (3) USA vs. Canada. For each pair, the model is given the same prompt asking it to predict the top-5 most likely countries for each image. The results are shown with color-coded annotations, where green indicates a correct top-1 prediction, red indicates an incorrect top-1, and yellow denotes cases where the ground truth appears within the top-5 but not at the top. The first pair, the model correctly identifies one country; in the second pair, it fails to correctly identify either; and in the third pair, it achieves one correct prediction and one partial success within the top-5. Overall, the examples demonstrate the model's difficulty in distinguishing visually similar neighboring regions.

\begin{table}[H]
\centering
\caption{
Neighbouring country confusion analysis using \textbf{Qwen3-VL-4B-Instruct}
in a blind two-turn setting (Turn~1: predict; Turn~2: justify).
\colorbox{correct}{Green} = correct Top-1,
\colorbox{wrong}{Red} = wrong Top-1,
\colorbox{partial}{Yellow} = GT in Top-5 but not Top-1.
}
\label{tab:neighbour_confusion}
\setlength{\tabcolsep}{4pt}
\renewcommand{\arraystretch}{1.2}
\begin{adjustbox}{max width=\textwidth}
\begin{tabular}{
>{\bfseries\raggedright}p{1.6cm}
m{3cm} m{3cm}
m{3cm} m{3cm}
m{3cm} m{3cm}
}
\toprule
& \multicolumn{2}{c}{\textbf{Pair 1: India / Pakistan}}
& \multicolumn{2}{c}{\textbf{Pair 2: India / Bangladesh}}
& \multicolumn{2}{c}{\textbf{Pair 3: USA / Canada}} \\
\cmidrule(lr){2-3}
\cmidrule(lr){4-5}
\cmidrule(lr){6-7}
& \textbf{Pakistan} & \textbf{India}
& \textbf{India} & \textbf{Bangladesh}
& \textbf{Canada} & \textbf{USA} \\
\midrule
Images &
\includegraphics[width=\linewidth,height=2.6cm,keepaspectratio]{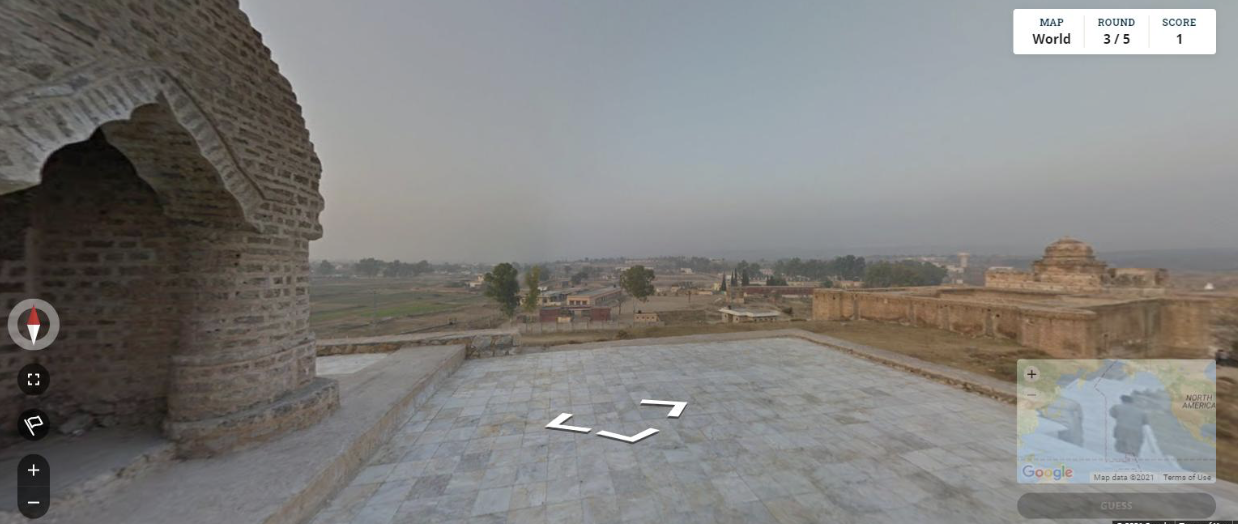} &
\includegraphics[width=\linewidth,height=2.6cm,keepaspectratio]{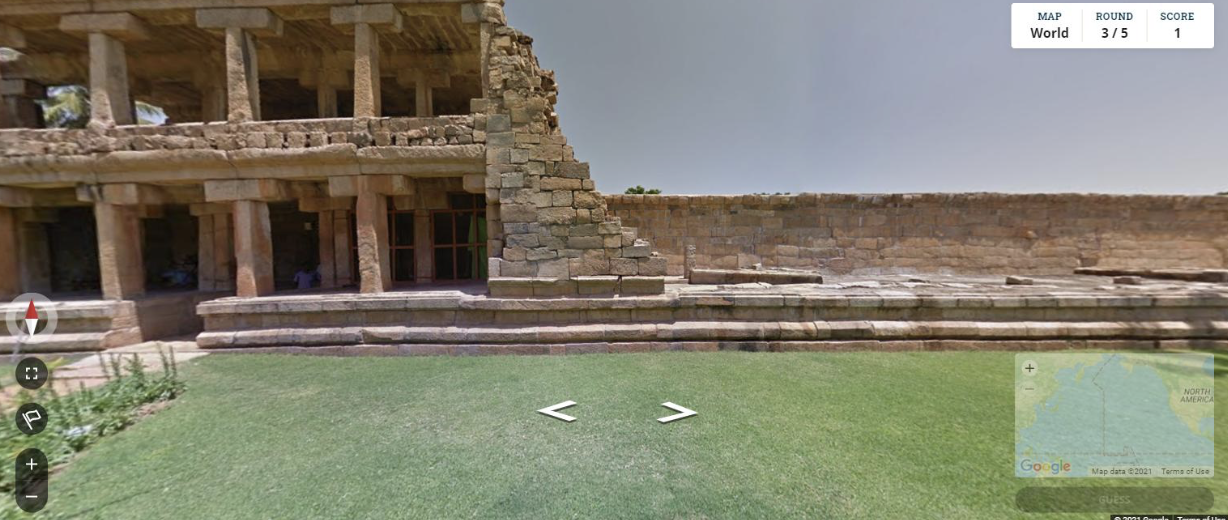} &
\includegraphics[width=\linewidth,height=2.6cm,keepaspectratio]{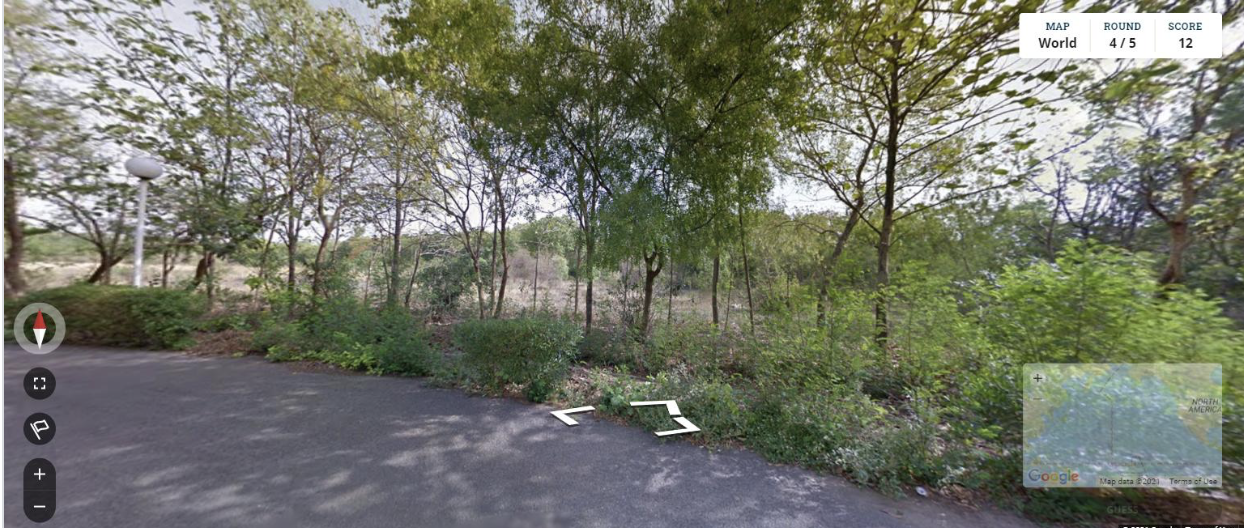} &
\includegraphics[width=\linewidth,height=2.6cm,keepaspectratio]{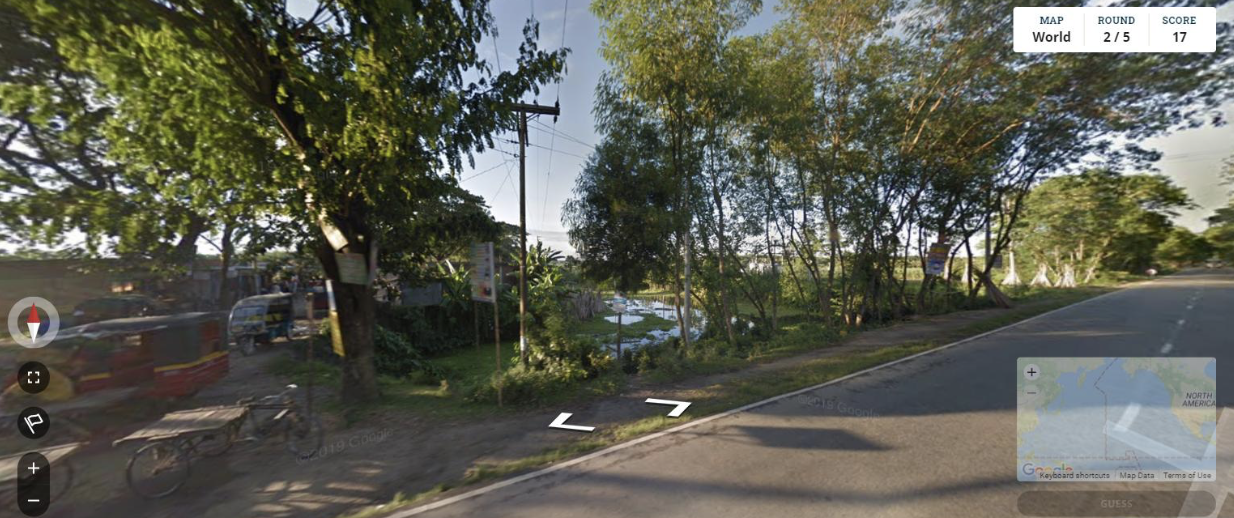} &
\includegraphics[width=\linewidth,height=2.6cm,keepaspectratio]{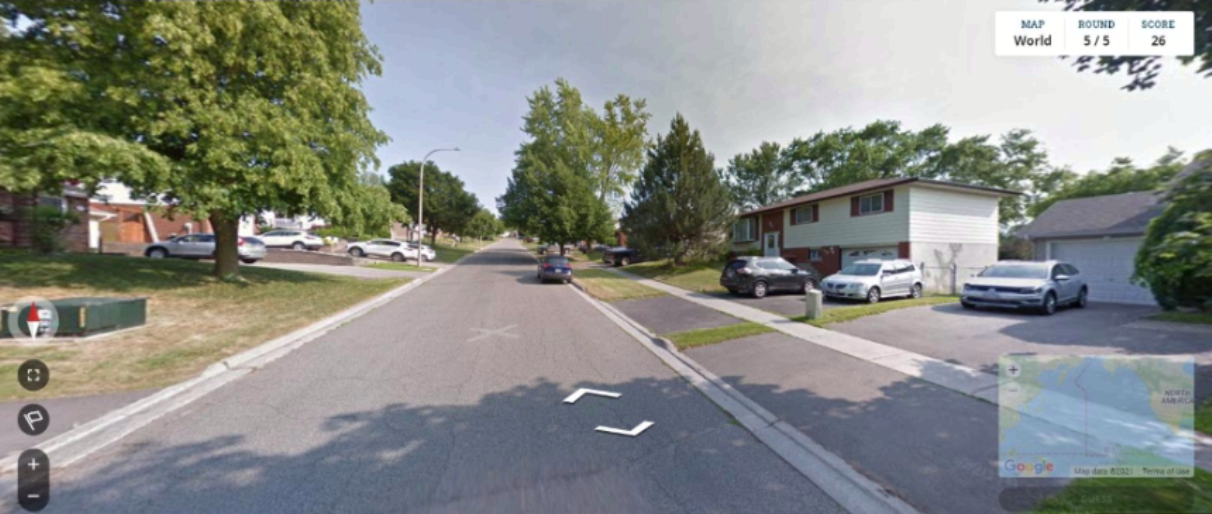} &
\includegraphics[width=\linewidth,height=2.6cm,keepaspectratio]{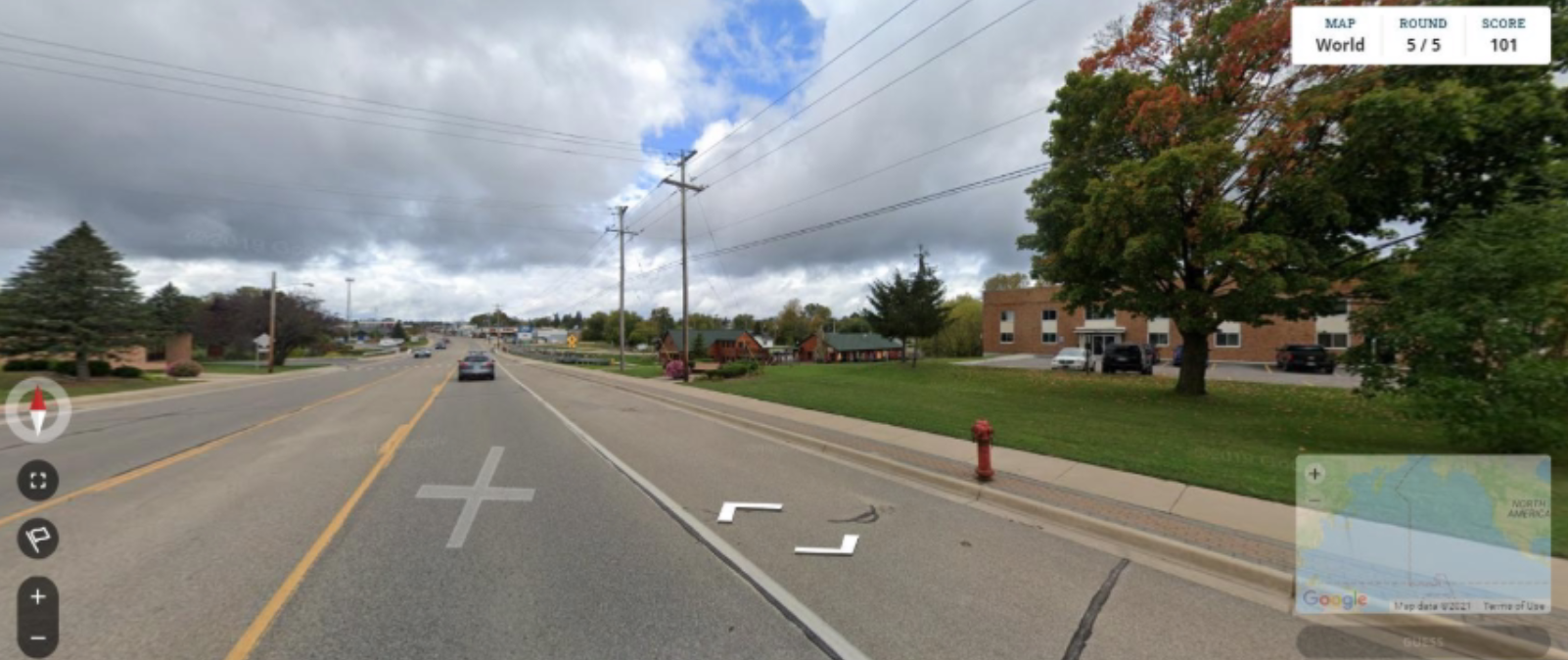} \\
\midrule
Prompt &
\multicolumn{2}{m{6.2cm}}{\small\textit{``I am showing you two images from two different locations in the world. For each image, predict the top 5 countries it is most likely from. Respond only in JSON format.''}} &
\multicolumn{2}{m{6.2cm}}{\small\textit{``I am showing you two images from two different locations in the world. For each image, predict the top 5 countries it is most likely from. Respond only in JSON format.''}} &
\multicolumn{2}{m{6.2cm}}{\small\textit{``I am showing you two images from two different locations in the world. For each image, predict the top 5 countries it is most likely from. Respond only in JSON format.''}} \\
\midrule
Predictions \& Justification
& \cellcolor{partial}\small
\textbf{Top-1: India~\xmark}\par
1.~India\par 2.~Pakistan\par 3.~United States\par 4.~Australia\par 5.~Canada\par
{\tiny\textit{Stone arch + arid landscape $\rightarrow$ defaulted to India.}}
& \cellcolor{correct}\small
\textbf{Top-1: India~\cmark}\par
1.~India\par 2.~Pakistan\par 3.~United States\par 4.~Australia\par 5.~Canada\par
{\tiny\textit{Ancient temple, stone fortification $\rightarrow$ India ranked \#1.}}
& \cellcolor{wrong}\small
\textbf{Top-1: USA~\xmark}\par
1.~United States\par 2.~Canada\par 3.~Mexico\par 4.~Brazil\par 5.~Argentina\par
{\tiny\textit{Street sign + American-style lamppost misled model.}}
& \cellcolor{wrong}\small
\textbf{Top-1: India~\xmark}\par
1.~India\par 2.~Indonesia\par 3.~Thailand\par 4.~Philippines\par 5.~Vietnam\par
{\tiny\textit{Dense tropical forest $\rightarrow$ South/SE Asia.}}
& \cellcolor{partial}\small
\textbf{Top-1: USA~\xmark}\par
1.~United States\par 2.~Canada\par 3.~Australia\par 4.~New Zealand\par 5.~Mexico\par
{\tiny\textit{Suburban homes + fire hydrant $\rightarrow$ defaulted to USA.}}
& \cellcolor{correct}\small
\textbf{Top-1: USA~\cmark}\par
1.~United States\par 2.~Canada\par 3.~Australia\par 4.~New Zealand\par 5.~Mexico\par
{\tiny\textit{Fire hydrant + utility poles $\rightarrow$ USA ranked \#1.}} \\
\bottomrule
\end{tabular}
\end{adjustbox}
\end{table}

\section{Biome Label Distribution}
\label{sec:suppl_biome_dist}


Table \ref{tab:suppl_biome_consensus_dist} summarizes the distribution of consensus biome labels across the three datasets, considering only the samples where all three annotators-CLIP, SigLIP, and Qwen3-VL-4B-are in agreement. It reports the frequency of each biome category under this strict consensus criterion, providing an overview of the recognized environmental classes across datasets.

\begin{table}[H]
\centering
\caption{Distribution of consensus biome labels where all three annotators (CLIP, Qwen, SigLIP) agree. CityGuessr-68k: 10{,}264 of 68{,}269 (15.0\%); GeoGuessr-50k: 13{,}400 of 49{,}997 (26.8\%); OSV5M: 18{,}001 of 50{,}000 (36.0\%).}
\label{tab:suppl_biome_consensus_dist}
\setlength{\tabcolsep}{3pt}
\renewcommand{\arraystretch}{1.22}
\begin{adjustbox}{max width=0.65\textwidth}
\begin{tabular}{l rr rr rr}
\toprule
& \multicolumn{2}{c}{\textbf{GeoGuessr-50k}} & \multicolumn{2}{c}{\textbf{CityGuessr-68k}} & \multicolumn{2}{c}{\textbf{OSV5M}} \\
\cmidrule(lr){2-3} \cmidrule(lr){4-5} \cmidrule(lr){6-7}
\textbf{Biome} & \textbf{Samples} & \textbf{\%} & \textbf{Samples} & \textbf{\%} & \textbf{Samples} & \textbf{\%} \\
\midrule
Temperate     & 8{,}446 & 63.0 & 1{,}609 & 15.7 & 5{,}910 & 32.8 \\
Mediterranean & 1{,}514 & 11.3 & 4{,}710 & 45.9 &     96  &  0.5 \\
Arid          & 1{,}646 & 12.3 & 3{,}115 & 30.3 & 8{,}304 & 46.1 \\
Boreal        & 1{,}607 & 12.0 &     99  &  1.0 &    569  &  3.2 \\
Tundra        &    109  &  0.8 &    683  &  6.7 &    785  &  4.4 \\
Tropical      &     78  &  0.6 &     48  &  0.5 & 2{,}337 & 13.0 \\
\bottomrule
\end{tabular}
\end{adjustbox}
\end{table}

\section{Biome Classification Prompts}
\label{sec:suppl_biome_prompts}

Table~\ref{tab:biome_prompts} lists the text prompts used for zero-shot biome classification. For CLIP and SigLIP, each prompt is used as a text embedding and the highest cosine similarity biome is assigned. For Qwen3-VL-4B, the six category names and descriptions are provided directly and the model is asked to select one.

\begin{table}[H]
\centering
\caption{Prompts used for zero-shot biome classification. CLIP and SigLIP use these as text embeddings for cosine similarity; Qwen3-VL-4B is prompted to select among these descriptions.}
\label{tab:biome_prompts}
\begin{tabular}{ll}
\toprule
\textbf{Biome} & \textbf{Prompt} \\
\midrule
Tropical      & ``a tropical rainforest or jungle scene'' \\
Arid          & ``a dry desert or arid landscape'' \\
Temperate     & ``a temperate forest or grassland scene'' \\
Mediterranean & ``a Mediterranean coastal or dry summer landscape'' \\
Tundra        & ``a cold tundra, snow, or polar landscape'' \\
Boreal        & ``a boreal forest or taiga with conifer trees'' \\
\bottomrule
\end{tabular}
\end{table}

\section{Top-5 Biome Accuracy}
\label{sec:suppl_biome_top5}

Table \ref{tab:biome_consensus_full} presents the Top-5 accuracy for nine Vision-Language Models (VLMs) across three datasets and six environmental biomes: Arid, Boreal, Mediterranean, Temperate, Tropical, and Tundra. The data highlights a dominant performance by the Qwen3-VL family, particularly the 2B and 4B variants, which achieve the highest scores in most categories. Conversely, the InternVL2.5 series shows accuracy steadily improving as model size increases. Finally, the results underscore that tropical regions are the most challenging to geolocalize across all models, whereas boreal regions often yield the highest accuracy due to more distinctive geographic cues.

\begin{table}[H]
\centering
\caption{Top-5 accuracy (\%) across biome categories using consensus
labels (all three annotators agree). GeoGuessr-50k: $n{=}$13{,}400 (26.8\%);
CityGuessr: $n{=}$10{,}264 (15.0\%); OSV5M: $n{=}$18{,}001 (36.0\%).
Best per-biome results are \textbf{bold}. [*L\,=\,LLaVA]}
\label{tab:biome_consensus_full}
\setlength{\tabcolsep}{4pt}
\renewcommand{\arraystretch}{1.15}
\begin{adjustbox}{max width=\textwidth}
\begin{tabular}{l cccccc cccccc cccccc}
\toprule
& \multicolumn{6}{c}{\textbf{GeoGuessr-50k} ($n{=}$13{,}400)}
& \multicolumn{6}{c}{\textbf{CityGuessr} ($n{=}$10{,}264)}
& \multicolumn{6}{c}{\textbf{OSV5M} ($n{=}$18{,}001)} \\
\cmidrule(lr){2-7} \cmidrule(lr){8-13} \cmidrule(lr){14-19}
Model & Arid & Bor. & Med. & Tmp. & Tro. & Tun.
      & Arid & Bor. & Med. & Tmp. & Tro. & Tun.
      & Arid & Bor. & Med. & Tmp. & Tro. & Tun. \\
\midrule
InternVL2.5-1B
  & 69.9 & 28.7 & 69.0 & 66.2 & 30.8 & 39.4
  & 64.7 & 52.5 & 83.0 & 60.4 & 54.2 & 60.5
  & 45.5 & 85.8 & 62.5 & 58.7 & 26.4 & 52.1 \\
InternVL2.5-4B
  & 67.6 & 71.4 & 76.4 & 73.0 & 38.5 & 45.0
  & 74.1 & 63.6 & 83.7 & 62.6 & 52.1 & 65.3
  & 45.4 & 90.7 & 61.5 & 58.5 & 26.4 & 47.6 \\
InternVL2.5-8B
  & 76.0 & \textbf{93.3} & 85.6 & 83.0 & 51.3 & 60.6
  & \textbf{82.2} & 65.7 & 84.1 & 66.4 & 58.3 & 70.7
  & 49.5 & 96.5 & 76.0 & 61.4 & 25.7 & 61.3 \\
\midrule
*L-Mistral-7B
  & 48.7 & 45.6 & 60.2 & 62.8 & 37.2 & 44.0
  & 63.8 & 15.2 & 55.3 & 47.0 & 31.2 & 40.3
  & 33.5 & 88.2 & 57.3 & 43.2 & 16.3 & 54.3 \\
*L-Vicuna-7B
  & 57.4 & 69.4 & 46.0 & 61.3 & 11.5 & 39.4
  & 55.2 & 46.5 & 57.6 & 52.5 & 37.5 & 40.4
  & 34.6 & 87.3 & 41.7 & 48.0 &  9.8 & 30.2 \\
*L3-LLaVA-8B
  & 61.8 & 87.7 & 69.2 & 69.5 & 43.6 & 48.6
  & 55.3 & 57.6 & 65.2 & 59.1 & 37.5 & 54.9
  & 42.1 & 96.0 & 54.2 & 53.2 & 21.1 & 52.6 \\
\midrule
Qwen3-VL-2B
  & \textbf{81.4} & 87.2 & 88.1 & 85.0 & 67.9 & 71.6
  & 78.2 & 64.6 & 87.5 & 71.1 & \textbf{60.4} & 71.7
  & 60.3 & 98.1 & \textbf{80.2} & \textbf{72.0} & \textbf{36.5} & 71.8 \\
Qwen3-VL-4B
  & 80.1 & 93.2 & \textbf{89.8} & \textbf{87.1} & \textbf{70.5} & \textbf{76.1}
  & 80.8 & 65.7 & \textbf{89.2} & \textbf{74.8} & \textbf{60.4} & 78.8
  & \textbf{61.6} & \textbf{98.4} & 78.1 & \textbf{72.0} & 33.8 & 70.6 \\
Qwen3-VL-8B
  & 66.9 & 72.1 & 83.8 & 77.8 & 64.1 & \textbf{76.1}
  & 81.5 & \textbf{66.7} & 88.9 & 74.6 & 54.2 & \textbf{83.6}
  & 57.7 & 97.0 & \textbf{80.2} & 69.8 & 32.0 & \textbf{75.5} \\
\bottomrule
\end{tabular}
\end{adjustbox}
\end{table}

\section{Preliminary LoRA Fine-Tuning}
\label{sec:suppl_lora}

To assess parameter-efficient fine-tuning, we fine-tune InternVL2.5-1B using LoRA ($r{=}8$, $\alpha{=}32$, all-linear modules) on a top-20 country subset of GeoGuessr-50k (12,116 train / 3,030 test images), training for 2 epochs at $5 \times 10^{-5}$. Top-1 accuracy improves from 40.40\% to \textbf{79.44\%} (+39.04\,pp). The gain is larger for rural scenes (+45.96\,pp) than urban (+25.70\,pp), suggesting the zero-shot model's rural weakness is a data-coverage issue addressable by fine-tuning. LoRA also improves error quality: GER-Weak rises from 17.55\% to 39.17\%, indicating the fine-tuned model's remaining errors are over twice as likely to be visually justified. Full exploration of larger ranks and scales is left to future work.

\section{Full Urban/Rural and GER Tables}
\label{sec:suppl_full_tables}
Table~\ref{tab:urban_rural_multi_full} extends Table 4 (main paper) with per-cell standard deviations. Urban accuracy is highly stable across labellers (std $<$ 1\,pp for most cells), while rural variance reaches 5--6\,pp, confirming that annotator disagreement concentrates on the rural boundary. Top-5 gaps are consistently narrower than Top-1, e.g., Qwen3-VL-4B's urban--rural difference on GeoGuessr-50k compresses from ${\sim}$15\,pp (Top-1) to ${\sim}$8\,pp (Top-5).

\begin{table}[H]
\centering
\caption{\textbf{Urban/Rural} Top-1 and Top-5 accuracy (\%), reported as mean $\pm$ std
across three independent labellers (CLIP, SigLIP, Qwen3-VL-4B).
The maximum standard deviation across all cells is 6.66\,pp.
[*L\,=\,LLaVA]}
\label{tab:urban_rural_multi_full}
\setlength{\tabcolsep}{2pt}
\renewcommand{\arraystretch}{1.1}
\begin{adjustbox}{max width=\textwidth}
\begin{tabular}{l cccc cccc cccc}
\toprule
& \multicolumn{4}{c}{\textbf{GeoGuessr-50k}}
& \multicolumn{4}{c}{\textbf{CityGuessr}}
& \multicolumn{4}{c}{\textbf{OSV5M}} \\
\cmidrule(lr){2-5} \cmidrule(lr){6-9} \cmidrule(lr){10-13}
& \multicolumn{2}{c}{Top-1} & \multicolumn{2}{c}{Top-5}
& \multicolumn{2}{c}{Top-1} & \multicolumn{2}{c}{Top-5}
& \multicolumn{2}{c}{Top-1} & \multicolumn{2}{c}{Top-5} \\
\cmidrule(lr){2-3} \cmidrule(lr){4-5}
\cmidrule(lr){6-7} \cmidrule(lr){8-9}
\cmidrule(lr){10-11} \cmidrule(lr){12-13}
Model
& Urb & Rur
& Urb & Rur
& Urb & Rur
& Urb & Rur
& Urb & Rur
& Urb & Rur \\
\midrule
InternVL2.5-1B
  & 48.81$\pm$.43 & 27.73$\pm$1.21
  & 76.14$\pm$.55 & 59.76$\pm$1.39
  & 46.93$\pm$.64 & 30.06$\pm$3.48
  & 68.07$\pm$.49 & 46.44$\pm$4.25
  & 33.57$\pm$5.96 & 30.15$\pm$.98
  & 54.69$\pm$6.55 & 51.22$\pm$1.28 \\
InternVL2.5-4B
  & 69.29$\pm$.85 & 46.71$\pm$1.57
  & 84.23$\pm$.54 & 67.54$\pm$1.25
  & 57.60$\pm$.71 & 34.89$\pm$4.46
  & 71.16$\pm$.53 & 50.20$\pm$4.19
  & 36.38$\pm$6.57 & 31.66$\pm$1.20
  & 54.35$\pm$6.39 & 50.25$\pm$1.16 \\
InternVL2.5-8B
  & 74.47$\pm$.34 & 58.35$\pm$1.34
  & 88.29$\pm$.23 & 78.67$\pm$.87
  & 59.61$\pm$.58 & 38.16$\pm$3.72
  & 72.74$\pm$.49 & 52.48$\pm$4.21
  & 38.24$\pm$6.08 & 36.26$\pm$1.23
  & 56.67$\pm$5.39 & 54.77$\pm$1.06 \\
\midrule
*L-Mistral-7B
  & 46.53$\pm$1.81 & 34.36$\pm$1.64
  & 59.72$\pm$1.41 & 49.43$\pm$1.26
  & 31.95$\pm$.61 & 19.16$\pm$4.55
  & 48.75$\pm$.58 & 28.31$\pm$4.62
  & 19.60$\pm$4.89 & 17.38$\pm$1.08
  & 33.93$\pm$5.06 & 36.02$\pm$1.19 \\
*L-Vicuna-7B
  & 42.76$\pm$2.11 & 30.73$\pm$1.82
  & 58.99$\pm$1.40 & 50.03$\pm$1.27
  & 38.35$\pm$.69 & 23.75$\pm$4.85
  & 53.16$\pm$.66 & 32.95$\pm$5.12
  & 22.26$\pm$5.00 & 18.82$\pm$1.09
  & 38.15$\pm$5.79 & 36.26$\pm$1.26 \\
*L3-LLaVA-8B
  & 59.85$\pm$.55 & 42.86$\pm$1.41
  & 74.89$\pm$.14 & 64.65$\pm$1.05
  & 40.73$\pm$.53 & 26.13$\pm$4.45
  & 58.75$\pm$.42 & 40.25$\pm$4.23
  & 28.53$\pm$5.17 & 28.00$\pm$1.16
  & 46.25$\pm$5.73 & 46.98$\pm$1.29 \\
\midrule
Qwen3-VL-2B
  & 82.56$\pm$.43 & 67.78$\pm$1.14
  & 92.04$\pm$.22 & 82.98$\pm$.78
  & 67.25$\pm$.69 & 43.27$\pm$4.77
  & 79.28$\pm$.46 & 61.07$\pm$2.51
  & 50.96$\pm$6.23 & 46.07$\pm$1.14
  & 68.68$\pm$4.55 & 64.01$\pm$.77 \\
Qwen3-VL-4B
  & 83.83$\pm$.49 & 68.49$\pm$1.16
  & 93.38$\pm$.18 & 85.03$\pm$.75
  & 68.84$\pm$.78 & 41.54$\pm$5.21
  & 80.10$\pm$.45 & 61.80$\pm$3.08
  & 51.26$\pm$6.30 & 44.80$\pm$1.06
  & 68.31$\pm$4.25 & 64.54$\pm$.79 \\
Qwen3-VL-8B
  & 79.82$\pm$.79 & 58.56$\pm$1.58
  & 89.50$\pm$.64 & 73.70$\pm$1.20
  & 68.31$\pm$.86 & 41.80$\pm$4.75
  & 80.29$\pm$.54 & 63.87$\pm$2.65
  & 48.23$\pm$6.66 & 41.05$\pm$1.10
  & 65.77$\pm$5.27 & 62.63$\pm$.98 \\
\bottomrule
\end{tabular}
\end{adjustbox}
\end{table}

Table~\ref{tab:ger_multi_full} supplements Table 4 (main paper) with per-cell standard deviations across the CLIP and SigLIP neighbour sets. The low maximum standard deviations (1.44\,pp for GER-W, 0.85\,pp for GER-S) confirm that GER rankings are robust to the choice of embedding model. Weaker models show higher relative variance (e.g., LLaVA-Mistral-7B: $13.25\pm1.44$ GER-W on GeoGuessr-50k), while the Qwen3-VL family is near-identical across both backbones (e.g., $17.18\pm0.04$ on OSV5M).

\begin{table}[H]
\centering
\caption{\textbf{Geographic Error Reasonableness (GER)} reported as mean $\pm$ std across CLIP and SigLIP nearest neighbours ($K{=}5$);
maximum SD across all cells is 1.44\,pp
(GER-W: 1.44, GER-S: 0.85).
GER-W: predicted country in $\geq$1 of 5 neighbours' ground-truth
countries; GER-S: in $\geq$2.
Computed over incorrect Top-1 predictions only (\%).
[*L\,=\,LLaVA]}
\label{tab:ger_multi_full}
\setlength{\tabcolsep}{4pt}
\renewcommand{\arraystretch}{1.1}
\begin{adjustbox}{max width=0.75\textwidth}
\begin{tabular}{l cc cc cc}
\toprule
& \multicolumn{2}{c}{\textbf{GeoGuessr-50k}}
& \multicolumn{2}{c}{\textbf{CityGuessr}}
& \multicolumn{2}{c}{\textbf{OSV5M}} \\
\cmidrule(lr){2-3} \cmidrule(lr){4-5} \cmidrule(lr){6-7}
Model & GER-W & GER-S & GER-W & GER-S & GER-W & GER-S \\
\midrule
InternVL2.5-1B  & 15.98$\pm$.53 & 7.18$\pm$.28  & 5.32$\pm$1.15 & 1.98$\pm$.43 & 11.52$\pm$.42 & 4.70$\pm$.15 \\
InternVL2.5-4B  & 21.34$\pm$1.27 & 11.43$\pm$.65 & 7.34$\pm$1.38 & 2.88$\pm$.50 & 11.54$\pm$.55 & 5.00$\pm$.25 \\
InternVL2.5-8B  & 33.72$\pm$.63 & 18.65$\pm$.30 & 7.34$\pm$1.29 & 2.94$\pm$.48 & 14.08$\pm$.32 & 6.22$\pm$.10 \\
\midrule
*L-Mistral-7B   & 13.25$\pm$1.44 & 6.97$\pm$.75  & 3.16$\pm$.76 & 1.17$\pm$.29 & 5.69$\pm$.39  & 2.42$\pm$.19 \\
*L-Vicuna-7B    & 10.67$\pm$1.25 & 5.70$\pm$.59  & 3.10$\pm$.67 & 1.18$\pm$.31 & 5.53$\pm$.39  & 2.45$\pm$.20 \\
*L3-LLaVA-8B    & 21.41$\pm$1.44 & 10.80$\pm$.85 & 3.95$\pm$.90 & 1.44$\pm$.32 & 10.02$\pm$.59 & 4.14$\pm$.28 \\
\midrule
Qwen3-VL-2B     & 40.48$\pm$.49 & 23.19$\pm$.47 & 7.84$\pm$.94  & 3.06$\pm$.29 & 17.18$\pm$.04 & 7.56$\pm$.03 \\
Qwen3-VL-4B     & \textbf{44.37$\pm$.45} & \textbf{25.92$\pm$.76} & \textbf{8.30$\pm$1.04} & \textbf{3.34$\pm$.43} & \textbf{17.14$\pm$.14} & \textbf{7.77$\pm$.02} \\
Qwen3-VL-8B     & 30.22$\pm$.58 & 16.96$\pm$.44 & 7.79$\pm$.95  & 3.11$\pm$.40 & 15.35$\pm$.19 & 6.73$\pm$.13 \\
\bottomrule
\end{tabular}
\end{adjustbox}
\end{table}

\end{document}